\documentclass[10pt]{article} 
\usepackage[accepted]{tmlr}

\usepackage{hyperref}
\usepackage{url}
\usepackage{graphicx}
\usepackage{multirow}

\usepackage{blindtext}
\usepackage{placeins}
\usepackage{enumerate}

\usepackage{xcolor}

\usepackage[utf8]{inputenc}
\usepackage[T1]{fontenc}
\usepackage{changepage}
\newcommand{\nystrom}{Nystr\"{o}m}
\newcommand{\erdosrenyi}{Erd\H{o}s-R\'enyi}

\usepackage{lastpage}

\title{Flexible Infinite-Width Graph Convolutional Neural Networks}

\author{\name Ben Anson \email ben.anson@bristol.ac.uk \\
        \addr School of Mathematics\\
        University of Bristol\\
        Bristol, United Kingdom\\
       \AND
        \name Edward Milsom \email edward.milsom@bristol.ac.uk \\
        \addr School of Mathematics\\
        University of Bristol\\
        Bristol, United Kingdom\\
       \AND
      \name Laurence Aitchison \email laurence.aitchison@bristol.ac.uk \\
       \addr School of Computer Science\\
       University of Bristol\\
       Bristol, United Kingdom\\}

\def\month{05}  
\def\year{2025} 
\def\openreview{\url{https://openreview.net/forum?id=Q2M4yijKSo}} 

\usepackage{booktabs}
\usepackage{amsmath,amsfonts,bm}
\newcommand{\E}{\mathbb{E}}

\newcommand{\KL}{\mathrm{D}_{\mathrm{KL}}}

\DeclareMathOperator{\Tr}{Tr}

\def\1{\bm{1}}

\renewcommand{\E}{\mathbb{E}}

\newcommand{\reals}{\mathbb{R}}

\newcommand{\hA}{\mathbf{\hat{A}}}

\newcommand{\normal}{\mathcal{N}}

{}

\newcommand{\const}[1]{\mathrm{const}}

\newcommand{\abs}[1]{\mathopen|#1\mathclose|}

\newcommand{\G}{\boldsymbol{\mathbf{G}}}
\newcommand{\Gtt}{{\v{G}}_{\mathrm{tt}}}
\newcommand{\Git}{{\v{G}}_{\mathrm{it}}}
\newcommand{\Gti}{{\v{G}}_{\mathrm{ti}}}
\newcommand{\Gii}{{\v{G}}_{\mathrm{ii}}}
\newcommand{\Ktt}{{\v{K}}_{\mathrm{tt}}}
\newcommand{\Kit}{{\v{K}}_{\mathrm{it}}}
\newcommand{\Kti}{{\v{K}}_{\mathrm{ti}}}
\newcommand{\Kii}{{\v{K}}_{\mathrm{ii}}}

\newcommand{\Att}{{\v{A}}_{\mathrm{tt}}}
\newcommand{\Ait}{{\v{A}}_{\mathrm{it}}}
\newcommand{\Ati}{{\v{A}}_{\mathrm{ti}}}
\newcommand{\Aii}{{\v{A}}_{\mathrm{ii}}}

\newcommand{\ppi}{P_{\mathrm{i}}}
\newcommand{\ppt}{P_{\mathrm{t}}}
\newcommand{\pp}{P}
\newcommand{\Fi}{{\v{F}}_{\mathrm{i}}}
\newcommand{\Ft}{{\v{F}}_{\mathrm{t}}}

\newcommand{\Hi}{{\v{H}}_{\mathrm{i}}}
\newcommand{\Ht}{{\v{H}}_{\mathrm{t}}}

\renewcommand{\v}[1]{\boldsymbol{\mathbf{#1}}}

\newcommand{\pr}{\P}
\newcommand{\Q}{\operatorname{Q}}
\renewcommand{\P}{\operatorname{P}}

\newcommand{\dof}{$\nu$}

\newcommand{\Ahat}{\hat{\v A}}

\newcommand{\cora}{Cora}
\newcommand{\citeseer}{Citeseer}
\newcommand{\pubmed}{Pubmed}
\newcommand{\squirrel}{Squirrel}
\newcommand{\chameleon}{Chameleon}
\newcommand{\minesweeper}{Minesweeper}
\newcommand{\romanempire}{Roman Empire}
\newcommand{\reddit}{Reddit}
\newcommand{\arxiv}{Arxiv}
\newcommand{\amazonratings}{Amazon Ratings}
\newcommand{\blogcatalog}{Blog Catalog}
\newcommand{\flickr}{Flickr}
\newcommand{\penn}{Penn94}
\newcommand{\tolokers}{Tolokers}

\usepackage{algorithm}
\usepackage{algorithmic}

\begin{document}

\maketitle
\begin{abstract}%
A common theoretical approach to understanding neural networks is to take an infinite-width limit, at which point the outputs become Gaussian process (GP) distributed.
This is known as a neural network Gaussian process (NNGP).
However, the NNGP kernel is fixed and tunable only through a small number of hyperparameters, thus eliminating the possibility of representation learning.
This contrasts with finite-width NNs, which are often believed to perform well because they are able to flexibly learn representations for the task at hand.
Thus, in simplifying NNs to make them theoretically tractable, NNGPs may eliminate precisely what makes them work well (representation learning).
This motivated us to understand whether representation learning is necessary in a range of graph tasks.
We develop a precise tool for this task, the graph convolutional deep kernel machine.
This is very similar to an NNGP, in that it is an infinite width limit and uses kernels, but comes with a ``knob'' to control the amount of flexibility and hence representation learning.
We found that representation learning gives noticeable performance improvements for heterophilous node classification tasks, but less so for homophilous node classification tasks.
\end{abstract}

\section{Introduction}
A fundamental theoretical method for analyzing neural networks involves taking an infinite-width limit of a randomly initialized network.
In this setting, the outputs converge to a Gaussian process (GP), and this GP is known as a neural network Gaussian process or NNGP~\citep{neal1996bayesian,lee2018deep,matthews2018gaussian}.
NNGPs have been used to study various different neural network architectures, ranging from fully-connected~\citep{lee2018deep} to convolutional~\citep{novak2018bayesian,garriga2018deep} and graph networks~\citep{walker2019graph,niu2023graph}.

However, it is important to understand whether these NNGP results apply in practical neural network settings.
One way of assessing the applicability of NNGP results is to look at performance.
Specifically, if infinite-width NNGPs perform comparably with finite-width NNs, we can reasonably claim that infinite-width NNGPs capture the ``essence'' of what makes finite-width NNs work so well.
In contrast, if finite-width NNs perform better, that would indicate that NNGPs are missing something.

Why might NNGPs underperform relative to NNs?
One key property of the infinite-width NNGP limit is that it eliminates representation learning: the kernel of the GP is fixed, and cannot be tuned except through a very small number of hyperparameters.
This fixed kernel makes it straightforward to analyze the behaviour of networks in the NNGP limit.
However, the elegant fixed kernel comes at a cost.
In particular, the top-layer representation/kernel is highly flexible in finite networks, and this flexibility is critical to the excellent performance of deep networks~\citep{bengio2013representation,lecun2015deep}.
This indicates that in some cases, the infinite-width NNGP limit can ``throw the baby out with the bathwater'': in trying to simplify the system to enable theoretical analysis, the NNGP limit can eliminate some of the most important properties of deep networks that lead to them performing well.

In the setting of convolutional networks for CIFAR-10, this is precisely what was found: namely, that infinite-width NNGPs underperform finite-width NNs~\citep{adlam2023kernel,garriga2018deep,shankar2020neural}, and it has been hypothesised that this difference arises due to the lack of representation learning in NNGP~\citep{aitchison2020why,mackay1998introduction}.

In this paper, we consider the question of NNGP performance in the Graph Convolutional Network (GCN) setting. Surprisingly, prior work has shown that the graph convolutional NNGP performs comparably to the GCN~\citep{niu2023graph}, perhaps suggesting that representation learning is not important in graph settings.

However, our work indicates that this picture is incomplete.
In particular, we show that while the graph convolutional NNGP is competitive on some datasets, it is much worse than GCNs on other datasets.
This would suggest a hypothesis that representation learning in graph tasks is dataset dependent, with some datasets needing representation learning, and others not.

Importantly, testing this hypothesis rigorously is difficult, as there are many differences between the infinite-width, fixed representation graph convolutional NNGP and the finite-width, flexible representation GCN, not just that the GCN has representation learning.
Perhaps the best approach to testing the hypothesis would be to develop a variant of the NNGP with representation learning, and to see how performance changed as we altered the amount of representation learning allowed by this new model.
However, this is not possible with the traditional NNGP framework. 
Instead, we need to turn to recently developed deep kernel machines (DKMs)~\citep{dkm23,milsom2023cdkm}.
DKMs, like NNGPs, are obtained via an infinite-width limit, and in practice work entirely in terms of kernels.
The key difference is that DKMs allow for representation learning, while NNGPs do not.
Specifically, DKMs have a tunable parameter, $\nu$, which controls the degree of representation learning.
For small values of $\nu$, DKM representations are highly flexible and learned from data.
In contrast, as $\nu \rightarrow \infty$, flexibility is eliminated, and the DKM becomes equivalent to the NNGP.
The graph convolutional DKM therefore forms a precise tool for studying the need for representation learning in graph tasks.

Using the graph convolutional DKM, we examined the need for representation learning in a variety of graph tasks.
Node classification tasks in particular exist on a spectrum from homophilous, where adjacent nodes in a graph tend to have similar labels, to heterophilous, where adjacent nodes tend to be dissimilar.
We found that homophilous tasks tended not to require representation learning, while heterophilous tasks did require representation learning.

Concretely, our contributions are as follows:
\begin{itemize}
  \item We develop a graph convolutional variant of deep kernel machines (Section~\ref{sec:methods}).
  \item We propose two scalable inducing-point approximation schemes for the graph convolutional DKM (Section~\ref{sec:ind_point_schemes}, Appendix~\ref{app:sec:inducing_point_appendix}).
  \item We analyze the graph convolutional DKM in the linear setting, and provide a closed-form solution for the learned representations (Section~\ref{sec:lineardkm}).
  \item By considering the performance of graph convolutional NNGPs relative to GCNs and by tuning $\nu$ in the graph convolutional DKM, we find that representation learning improves performance in heterophilous node classification tasks but not homophilous node classification tasks.
\end{itemize}
\section{Related Work}
Graph convolutional networks are a type of graph neural network~\citep{scarselli2008graph,kipf2016semi,bronstein2017geometric,velivckovic2017graph} originally introduced by \citet{kipf2016semi}, motivated by a localised first-order approximation of spectral graph convolutions. They use the adjacency matrix to aggregate information from a node's neighbours at each layer of the neural network. When first published, graph convolutional networks outperformed existing approaches for semi-supervised node classification by a significant margin.

Homophily, a measure of similarity between connected nodes in a graph, has been studied extensively in the context of GNNs~\citep{luan2024graph,zhu2020beyond,zhu2021graph,maurya2021improving,platonov2023critical}, mostly to point out that lack of homophily (meaning that connected nodes tend to be dissimilar)  is detrimental to performance in GNNs. Here, we examine an adjacent but different matter: the relationship between homophily and representation learning.

Infinite-width neural networks were first considered in the 1990s~\citep{neal1996bayesian,williams1996computing} where they only considered shallow networks at initialisation.~\citet{cho2009kernel} derived an equivalence between deep ReLU networks and arccosine kernels. This was later generalised to arbitrary activation functions by~\citet{lee2018deep}, who used a Bayesian framework to show that any infinitely-wide deep neural network is equivalent to a Gaussian process, which they dubbed the neural network Gaussian process (NNGP)~\citep{lee2020finite,lee2018deep,matthews2018gaussian}. The NNGP has since been extended, for example to accommodate convolutions~\citep{garriga2018deep,novak2018bayesian}, graph structure~\citep{niu2023graph,hu2020infinitely} and more~\citep{yang2019scaling}.

Using a similar infinite-width limit, but on SGD-trained networks rather than Bayesian networks, \citet{jacot2018neural} studied the dynamics of neural network training under gradient descent, deriving the neural tangent kernel (NTK), which describes the evolution of the infinite-width network over time. The NTK suffers from an analogous problem to the NNGP's lack of representation learning, since the NTK limit implies that the intermediate features do not evolve over time~\citep{yang2021tensor,bordelon2023self,vyas2023feature}.  Alternative limits, such as the recent $\mu$-P parameterisation~\citep{yang2021tensor} fix this problem by altering the scaling of parameters.  However, the $\mu$-P line of work only tells us that feature/representation learning will happen, not what the learned features/representations will be. In contrast the DKM framework gives us the exact learned representations.

Applying {\it fixed\/} kernel functions to graphs is a well explored endeavour~\citep{shervashidze2009fast,kashima2003marginalized,shervashidze2009efficient}. Kernels can be defined directly for graphs and applied in a shallow fashion, shallow kernel can be stacked~\citep{achten2023semi}, or kernels can be used as a component of a larger deep learning algorithm~\citep{cosmo2021graph,yanardag2015deep}. Our work differs fundamentally in that kernels themselves are learned, rather than learning features and then applying a fixed kernel.
\section{Background}\label{sec:background}
We give a background on deep Gaussian processes (DGPs), and how under an infinite-width limit they become neural network Gaussian processes (NNGPs). We review how modifying this infinite-width limit gives flexible DKMs, in contrast to NNGPs which have fixed kernels.
We also give an overview of the graph convolutional network (GCN) and the graph convolutional NNGP. These ingredients allow us to define a DKM in the graph domain, a so-called ``graph convolutional DKM'', in Section~\ref{sec:methods}.

\subsection{Neural Networks as Deep Gaussian Processes}\label{sec:nn_as_dgp}
Consider a fully-connected NN with inputs $\v X\in\mathbb{R}^{P\times \nu_0}$, outputs/labels $\v Y\in\mathbb{R}^{P\times \nu_{L+1}}$, and intermediate-layer features $\mathbf{F}^{\ell}\in\reals^{P\times N_\ell}$.  By $P$ we refer to the number of datapoints, $\nu_0$ and $\nu_{L+1}$ the number of input and output features respectively.  The width of each layer is denoted $N_\ell = N \nu_\ell$, where $\nu_\ell$ is fixed, and $N$ allows us to scale the width of all layers simultaneously.
We compute features at each layer using the previous layer via $\mathbf{F}^{\ell} = \phi(\mathbf{F}^{\ell-1})\v W^\ell$, where $\v W^\ell \in \reals^{N_{\ell-1} \times N_\ell}$ are the weights and $\phi$ is the pointwise nonlinearity (e.g. ReLU).
With a Gaussian prior on the weights, $W_{ij}^\ell \sim \mathcal N(0,\tfrac{1}{N_{\ell - 1}})$, the conditional distribution of features is given by,
\begin{subequations}\label{eq:dgp}
\begin{align}
\pr(\v F^\ell \mid \v F^{\ell -1}) &= \prod_{\lambda=1}^{N_{\ell}}\normal(\v f_\lambda^\ell; \v 0, \v K_\text{features}(\v F^{\ell -1})),\\
\pr(\v Y\mid \v F^L)&=\prod^{\nu_{L+1}}_{\lambda=1}\normal(\v y_\lambda ; \v 0, \v K_\text{features}(\v F^L) + \sigma^2 \v I),
\end{align}
\end{subequations}
where $\v K_\text{features}(\v X) = \tfrac{1}{N_\ell}\phi(\v X)\phi(\v X)^T$, and the subscript $\lambda$ refers to the $\lambda$th feature/column, so that ${\v F^\ell = (\v f_1^\ell \,\, \v f_2^\ell \,\, \cdots \,\, \v f_{N_\ell}^\ell)}$.
Eq.~\eqref{eq:dgp} is equivalent to a Bayesian NN, but in the above form it is usually called a deep Gaussian process (DGP) \citep{damianou2013deep,salimbeni2017doubly}, albeit with an unusual kernel. We can generalize Eq.~\eqref{eq:dgp} to allow alternative kernels $\v K_\text{features}(\cdot)$ such as the squared exponential kernel.
\subsection{Deep Gaussian Processes in Terms of Gram Matrices}\label{sec:dgps_in_terms_of_grams}
For certain kernel functions (e.g.\ arccosine kernels~\citep{cho2009kernel} and RBF), features are unnecessary, and inter-layer dependencies in a DGP can be summarised entirely by Gram matrices ${\mathbf G^\ell = \tfrac{1}{N_\ell} \v F^{\ell} (\v F^{\ell})^T \in\mathbb{R}^{P\times P}}$.
For these kernels, we are able to compute the kernel from the Gram matrix alone, and thus there is a function over Gram matrices $\v K(\cdot) : \mathbb R^{P \times P} \to \mathbb R^{P \times P}$ such that  $\v K_\text{features}(\v F^{\ell})=\v K(\v G^\ell)$
\footnote{
For avoidance of doubt, ${\v K_\text{features}(\cdot) : \mathbb R^{P \times N_\ell} \to \mathbb R^{P \times P}}$ computes the kernel from the features, while ${\v K(\cdot) : \mathbb R^{P \times P} \to \mathbb R^{P \times P}}$ equivalently computes the same kernel from the corresponding Gram matrix.}.
We can therefore rewrite the DGP in Eq.~\eqref{eq:dgp} as,
\begin{subequations}\label{eq:dgp:G}
\begin{align}
\pr(\v F^\ell \mid \v G^{\ell -1}) &= \prod_{\lambda=1}^{N_{\ell}}\normal(\v f_\lambda^\ell; \v 0, \v K(\v G^{\ell-1})),\\
\pr(\v Y\mid \v G^L)&=\prod^{\nu_{L+1}}_{\lambda=1}\normal(\v y_\lambda ; \v 0, \v K(\v G^L) + \sigma^2 \v I),
\end{align}
\end{subequations}
where all inter-layer dependencies in Eq.~\eqref{eq:dgp:G} are expressed using only Gram matrices. We can remove features altogether, and reframe the model purely in terms of Gram matrices~\citep{aitchison2021deep},
\begin{subequations}\label{eq:dwp:G}
\begin{align}
\pr(\v G^\ell \mid \v G^{\ell -1}) &= \mathcal{W}(\v G^\ell; \v G^{\ell-1} / N_\ell, N_\ell),\\
\pr(\v Y\mid \v G^L)&=\prod^{\nu_{L+1}}_{\lambda=1}\normal(\v y_\lambda ; \v 0, \v K(\v G^L) + \sigma^2 \v I),
\end{align}
\end{subequations}
where $\mathcal{W}(\v V, n)$ is the Wishart distribution with scale matrix $\mathbf V$ and degrees of freedom $n$~\citep{gupta2018matrix}.
\subsection{Wide Deep Gaussian Processes}
Consider a deep ReLU network. Bayesian inference over weights/features is not tractable, but remarkably, there {\it is\/} a tractable closed form solution if we take an infinite width limit. To see this, first observe that we can calculate the kernel non-linearity without features;
\citet{cho2009kernel} showed that in the limit $N_\ell\rightarrow\infty$, ReLU kernels are equivalent to arccosine kernels:
\begin{align}\label{eq:k_features}
  \phi=\mathrm{ReLU}\implies  \lim_{N_\ell \to \infty} \tfrac{1}{N_\ell} \phi(\v F^{\ell}) \phi^T(\v F^{\ell})  &= \v K_\text{arccos}(\v G^{\ell}).
\end{align}
Further, the arccosine kernel is easily computable,
\begin{align}
K_\text{arccos}(\v G)_{ij} = \frac{\sqrt{G_{ii} G_{jj}}}{\pi}\left(\sin \theta_{ij} + (\pi - \theta_{ij})\cos\theta_{ij}\right),\text{ where }\cos\theta_{ij} = \frac{G_{ij}}{\sqrt{G_{ii} G_{jj}}}.
\end{align}
Secondly, Gram matrices become deterministic in wide networks, since they can be understood as an average of IID features,
$\v G^\ell = \tfrac{1}{N_\ell} {\textstyle\sum}_{\lambda=1}^{N_\ell} \v f^{\ell}_\lambda (\v f^{\ell}_\lambda)^T$.
By Eq.~\eqref{eq:dgp} or Eq.~\eqref{eq:dgp:G}, the expectation of $\v G^\ell$ conditioned on the previous layer is,
$\E[\v G^\ell | \v G^{\ell-1}] = \v K(\v G^{\ell-1})$, while the variance scales with $1/N_\ell$.
Thus, as we take an infinite-width limit, $N_\ell \rightarrow \infty$, the variance goes to zero, and the prior over the Gram matrix at layer $\ell$, $\v G^\ell$, becomes deterministic and concentrated at its expectation, $\v K(\v G^{\ell-1})$.
Thus, the kernel at every layer is a fixed deterministic function of the inputs that can be computed recursively,
\begin{align}
\label{eq:nngp}
    \v K_\text{network}(\v F^\ell) &= \underbrace{(\mathbf{K} \circ \cdots \circ \mathbf{K})}_{\ell\text{ times}}(\tfrac{1}{\nu_0} \mathbf{X}\mathbf{X}^T),
\end{align}
where $\tfrac{1}{\nu_0} \mathbf{X}\mathbf{X}^T$ is the kernel for the inputs, and $\mathbf F^0 = \mathbf X$.
Setting $\ell=L$, we see that Eq.~\eqref{eq:nngp} holds true at the output layer, implying that the outputs of a NN or DGP are GP distributed with a fixed kernel in the infinite-width limit. This is known as the neural network Gaussian process (NNGP)~\citep{lee2018deep}.
\subsection{Deep Kernel Machines}\label{sec:dkm}
The clean analytic form of the NNGP kernel (Eq.~\ref{eq:nngp}) unfortunately has a big weakness: it underperforms finite-width counterparts empirically in many tasks~\citep{aitchison2020why,pleiss2021limitations}. The NNGP kernel is a fixed function of inputs, meaning that no representation learning can occur, and the kernel cannot be shaped by the output labels. This is a big problem, since representation learning is widely understood to be central to the success of modern deep learning systems.

DKMs~\citep{dkm23,milsom2023cdkm} are a solution to the lack of representation learning in infinite-width NNGPs. DKMs can still be understood as an infinite-width limit of a DGP/NN, but the limit is altered to retain flexibility and allow learning of representations.
In a DKM, Gram matrices are treated as learned parameters, rather than random variables as in a DGP or NNGP.~\cite{dkm23} derive the following ``DKM objective'' for fully-connected networks that can be optimized with respect to Gram matrices $\v G^\ell$,
\begin{align}
\label{eq:dkm_obj}
\mathcal{L}(\v G^1,\ldots,\v G^L) = \log \P(\v Y \mid \v G^L)-\sum_{\ell=1}^L\nu_\ell \KL\left(\normal(\v 0, \v G^\ell)\mid\mid\normal(\v 0, \v  K(\v G^{\ell-1}))\right).
\end{align}
Here, the likelihood term $\P(\v Y \mid \v G^L)$ measures how well the final layer representation $\v G^L$ performs at the task. Under the derivation of~\cite{dkm23}, $\P(\v Y \mid \v G^L)$ is a likelihood for a Gaussian process distribution over the outputs with kernel $\v K(\v G^L)$. The KL divergence terms are a measure of how much the Gram matrices $\G^1,\,\ldots,\,\G^L$ deviate from the NNGP. This can be seen due to the fact that a negative KL divergence is maximized exactly when the two distributions it is measuring are equal, i.e.\ $\v G^\ell = \v K(\v G^{\ell-1})$.
 The DKM objective trades off the marginal likelihood against the KL divergence terms, with the KL terms acting as a regularizer towards the NNGP. The amount of regularization depends on the $\nu_\ell$ coefficients, and as we send $\nu_\ell \rightarrow \infty$, the DKM becomes equivalent to the standard infinite-width NNGP.
As $\nu_\ell$ becomes smaller, we allow more flexibility in the Gram representations.
Thus, through $\nu_\ell$, the DKM gives us a knob to tune the amount of representation learning allowed in our model.
\subsection{Graph Convolutional Networks and their Wide counterparts}
\citet{kipf2016semi} derived Graph Convolutional Networks (GCNs) by applying approximations to spectral methods on graphs. In a GCN layer, an adjacency matrix $\v A\in\pp\ \times \pp$ is used to convolve (or `mixup') intermediate layer nodes. Mathematically, GCN post-activations are given by $\v H^\ell = \phi(\hA\v H^{\ell-1}\v W^\ell)$, where $\hA$ is a normalized version of $\v A$.

Similarly to Section~\ref{sec:dgps_in_terms_of_grams}, it is possible to construct an NNGP for the graph domain.
The pre-activations ${\v F}^\ell$ of a graph convolution layer are $\mathbf{F}^\ell = \hat{\mathbf{A}} \mathbf{H}^{\ell-1}\mathbf{W}^\ell$.
Again, placing an IID Gaussian prior over our weights $W^\ell_{\mu\lambda} \sim \mathcal N(0,\tfrac{1}{N_{\ell-1}})$, the NNGP construction tells us that as we make the layers wide ($N_{\ell} \to \infty$), the pre-activations become GP distributed with kernel matrix
$\mathbf{K} = \mathbb E[\mathbf{f}^\ell_\lambda(\mathbf{f}^\ell_\lambda)^T]$. We show in Appendix~\ref{app:kernelderiv} that this expectation has closed-form,
\begin{align}\label{eq:gcnngp_kernelmatrix}
    \mathbf{K} = \hat{\mathbf{A}}\boldsymbol{\Phi}^{\ell-1}\hat{\mathbf{A}}^T,
\end{align}
where $\boldsymbol{\Phi}^{\ell-1} = \E[\mathbf{h}_\lambda^{\ell-1}(\mathbf{h}^{\ell-1}_\lambda)^T]$ is the NNGP kernel of a fully connected network (e.g. the arccosine kernel in the case of a ReLU network).
Using this kernel in the NNGP defined in Eq. \ref{eq:nngp} gives the graph convolutional NNGP recursion,
\begin{equation}\label{eq:exposition_gcnngp}
\G^\ell = \hat{\mathbf{A}}\mathbf{K}(\G^{\ell-1})\hat{\mathbf{A}}^T = \hat{\mathbf{A}}\mathbf{K}(\hat{\mathbf{A}}\mathbf{K}(\cdots \hA \v K(\tfrac{1}{\nu_0} \mathbf{X}\mathbf{X}^T) \hA^T\cdots)\hA^T)\hat{\mathbf{A}}^T.
\end{equation}
GCNs and graph convolutional NNGPs can be applied easily to node classification problems, since features or nodes are modelled directly. They can also be used for graph classification by applying mean pooling at the output layer.

In Sections~\ref{sec:lineardkm} and~\ref{sec:experiments} we consider homophily in graphs and its effect on graph convolutional NNGP performance. By homophily, we refer to edge homophily $h\in[0, 1]$, which is calculated as the proportion of edges $(j,k)\in\mathcal{E}$ that are between nodes with the same class.
\section{Methods}\label{sec:methods}
In Section~\ref{sec:background} we have seen how graph convolutional NNGP kernels can be calculated. We have also seen how DKMs add flexibility to fully-connected NNGPs. We now combine these two ideas to obtain a flexible infinite-width graph network --- the `graph convolutional DKM' --- and develop an inducing point scheme to allow training on large datasets.

\subsection{Graph Convolutional Deep Kernel Machines}\label{sec:gcdkms}
The graph convolutional NNGP suffers from the ``fixed representation'' problem
that plagues all NNGP models — its kernel is a fixed transformation of the
inputs, regardless of the target labels. This means that there is little
flexibility (apart from kernel hyperparameters) to learn suitable
features for the task at hand. To solve this problem we develop the graph
convolutional DKM,  which has learnable kernel representations at each layer.

We arrive at a graph convolutional DKM by considering the Gram matrices in a DGP with graph mixups,
\begin{align}
\pr(\v F^\ell \mid \v F^{\ell -1}) &= \prod_{\lambda=1}^{N_{\ell}}\normal(\v f_\lambda^\ell; \v 0, \hA \v K(\v G^{\ell-1})\hA^T),\label{eq:gdgp1}\\
\text{where }\v G^{\ell-1} &= \tfrac{1}{N_{\ell-1}}\v F^{\ell-1}(\v F^{\ell-1})^T.\nonumber
\end{align}
By taking the representation learning limit of the graph convolutional DGP, we show in Appendix~\ref{app:gcdkm_deriv} that the posterior over Gram matrices is point distributed; in other words, the Gram matrices are deterministic. Moreover, the Gram matrices maximize the graph DKM objective,
\begin{align}
\label{eq:gcdkm_obj}
\mathcal{L}(\v G^1,\ldots,\v G^L) = \log \P(\v Y \mid \v G^L)-\sum_{\ell=1}^L\nu_\ell \KL\left(\normal(\v 0, \v G^\ell)\mid\mid\normal(\v 0, \hA\v  K(\v G^{\ell-1}) \hA^T)\right).
\end{align}
The objective in Eq.~\eqref{eq:gcdkm_obj} can be interpreted as a fully-connected DKM, but with a modified kernel. Again, just as in Section~\ref{sec:dkm}, we trade-off the likelihood with the KL divergence terms. That is, we trade-off the suitability of the top-layer representation for explaining the labels $\v Y$ versus divergence from the graph convolutional NNGP. When $\nu_\ell\rightarrow\infty$ in the DKM objective (Eq.~\ref{eq:dkm_obj}) we recover the NNGP, and similarly for the graph DKM objective taking $\nu_\ell\rightarrow\infty$ gives us the graph convolutional NNGP. Decreasing the coefficients $\nu_\ell$ decreases the importance of the KL divergence terms, giving Gram matrices flexibility to deviate from the graph convolutional NNGP.
 In our experiments, we treat the $\nu_\ell$ as knobs for controlling the amount of representation learning. By measuring performance for different $\nu_\ell$, we are able to cleanly demonstrate the importance of learned representations in various tasks.

\subsection{Inducing-point schemes}\label{sec:ind_point_schemes}
In practice it is computationally infeasible to optimize the Gram matrices $\v G^1,\ldots,\v G^L$ in the graph DKM objective for anything but modestly sized datasets, due to quadratic memory costs of storing the Gram matrices $\v G^\ell$, and the cubic time costs. The KL divergence terms, for example, are expensive because of matrix inversions and log determinants:
\begin{align}
  \KL\big(\normal(\v 0,& \v G^\ell)\mid\mid\normal(\v 0, \hA{\v K}(\G^{\ell-1}) \hA^T)\big) =\nonumber\\
  \tfrac{1}{2}\big(&\mathrm{Tr}((\hA\v K(\v G^{\ell-1}) \hA^T)^{-1}\v G^{\ell }) -\log\det((\hA{\v K}(\v G^{\ell-1})\hA^T)^{-1} \G^{\ell })  - P\big).
\end{align}
We resolve computational issues by developing an inducing point scheme and enabling linear scaling with dataset size. 
Specifically, we suppose that the graph nodes $\v X\in\mathbb{R}^{P\times \nu_0}$ can be summarized sufficiently well by a smaller inducing set $\v X_\mathrm{i}\in\mathbb{R}^{\ppi\times \nu_0}$, where the size of the inducing set $\ppi$ is fixed and chosen to be much smaller than the full set of nodes (so that $\ppi \ll P$). These inducing inputs form an initial Gram matrix that we call $\Gii^0 = \tfrac{1}{\nu_0}\v X_\mathrm{i} \v X_\mathrm{i}^T\in\mathbb{R}^{\ppi\times \ppi}$. We also have inducing Gram representations at each layer, $\Gii^1,\ldots,\Gii^L$, with all of these inducing representations being learned. The inducing representations are propagated alongside training/test Gram representations $\Gtt^1,\ldots,\Gtt^L$, so that,
\begin{align}
  \G^\ell = \begin{pmatrix}
    \Gii^\ell & \Git^\ell\\
    \Gti^\ell & \Gtt^\ell
  \end{pmatrix}.
\end{align}
The inducing and training/test points interact via the similarity matrix $\Gti^\ell = (\Git^\ell)^T\in \mathbb{R}^{\ppt\times \ppi}$, and $\Gti^\ell$ and $\Gtt^\ell$ are predicted according to our inducing point scheme, in a similar way to how training/test features are predicted in an inducing point GP. We refer readers to Algorithm~\ref{alg:pred} for computational details.

The form of the inducing Gram matrices depends on the choice of inducing point scheme. The modeller must make a choice as to how the inducing graph nodes interact with the training/test nodes.
Broadly, we can either consider intra-domain inducing points, which belong to the same domain as the data, or inter-domain inducing points which do not.
We consider one scheme of each type: an intra-domain scheme that assumes adjacency information between inducing points/nodes and training/test nodes, and an inducing scheme that treats the inducing points as unconnected nodes in a graph.

We `connect' the inducing nodes to each other with an inducing adjacency matrix  $\Aii\in\mathbb{R}^{\ppi\times \ppi}$, and `connect' the inducing nodes to the training/test nodes with the adjacency $\Ati\in\mathbb{R}^{\ppt \times \ppi}$.
We assume that the adjacency between train/test points is provided in the dataset, which we denote $\Att$ for consistency. The two schemes are characterized by,
\begin{enumerate}[(1)]
\item (intra-domain) sampling a random subset $\mathcal{S} = \{s_1,\ldots,s_{\ppi}\}$ of nodes from the dataset and treating these as inducing points. The inducing adjacencies become $(\Aii)_{jk} = A_{s_j,s_k}$, $(\Ati)_{jk} = A_{j,s_k}$;
\item (inter-domain) treating inducing nodes as independent of all other nodes, such that $\Aii = \v I$, $\Ati = \v 0$.
\end{enumerate}
We give forward propagation rules for each inducing-point scheme, and an ELBO objective (which is used for optimizing the inducing points and replaces the full-rank objective function from Eq.~\ref{eq:gcdkm_obj}) in Appendix~\ref{sec:graph_ind_point_scheme}. The computational cost of the intra-domain scheme is slightly higher, due to having to store $\Aii$ and $\Ati$ and perform matrix multiplications with them. However this cost is negligible compared to storing and computing with the Gram matrices themselves, and further mitigated by the fact that the adjacencies are sparse.

We empirically compare both schemes in Appendix~\ref{app:ind_point_abl} across several node classification dataset, finding that neither is definitively the best. However, for some datasets (e.g.~\citeseer) the inter-domain scheme is markedly better. Given that intra-domain inducing points are not applicable in a multi-graph setting, and that the inter-domain scheme has a simpler implementation, the inter-domain is preferable overall.
\begin{figure*}[t]
    \centering
    \includegraphics[width=0.8\linewidth]{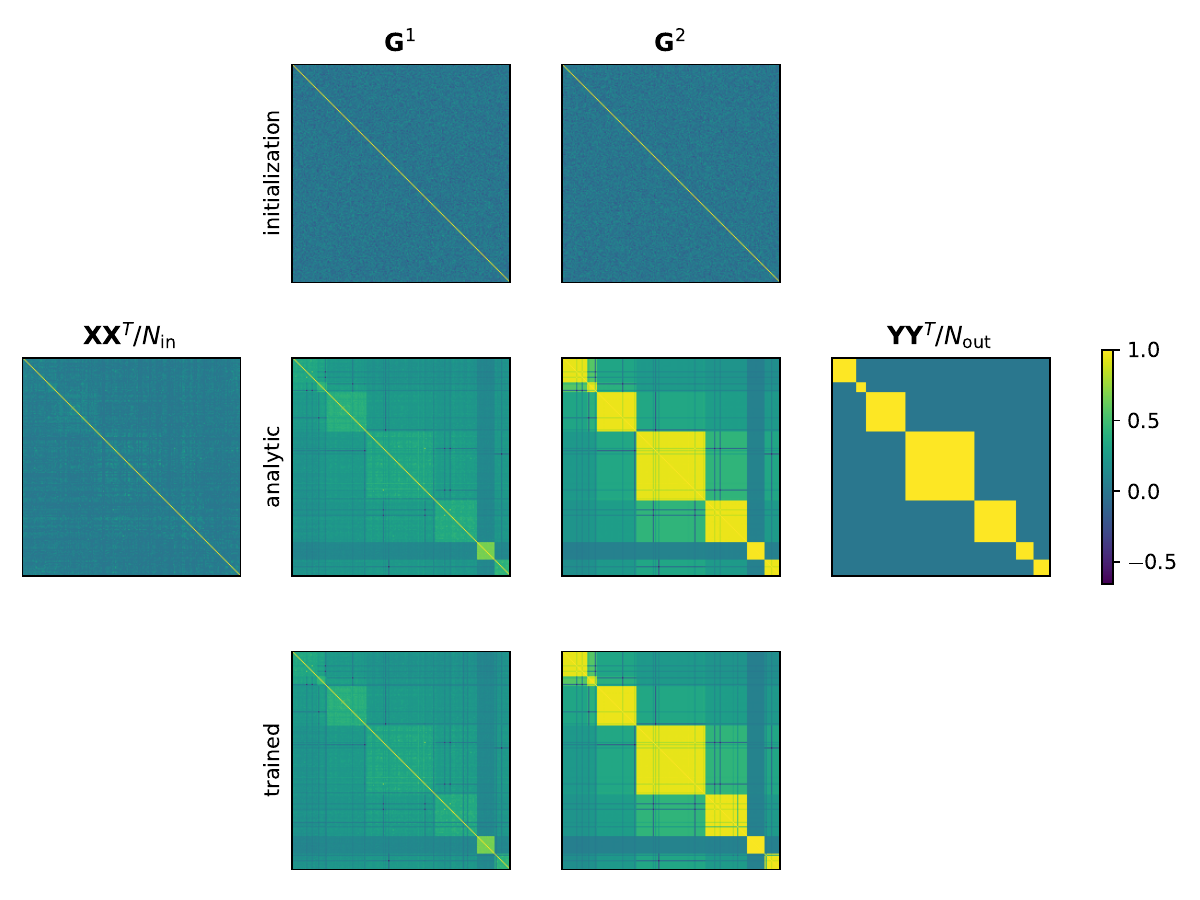}
    \caption{Normalized Gram matrices for a 2-layer linear graph DKM, providing verification of Eq.~\eqref{eq:linear_gcdkm_solution}. We use input data, labels, and adjacency from a randomly selected 100 datapoint subset of~\cora, and set the adjacency to be $\hA_{0.5}$. The leftmost and rightmost kernels are the input and label kernels respectively. The middle two columns are the hidden Gram representations. The top row shows Gram matrices after a Wishart initialization, the middle row shows the analytic solution, and the bottom row shows the results after training from this initialization by gradient descent with Adam.
    }
    \label{fig:linear_confirm_kernel}
\end{figure*}

\section{Analysis of Representation Learning in Linear Graph Convolutional Deep Kernel Machines}\label{sec:lineardkm}
\begin{figure*}[t]
    \centering
    \includegraphics[width=\textwidth]{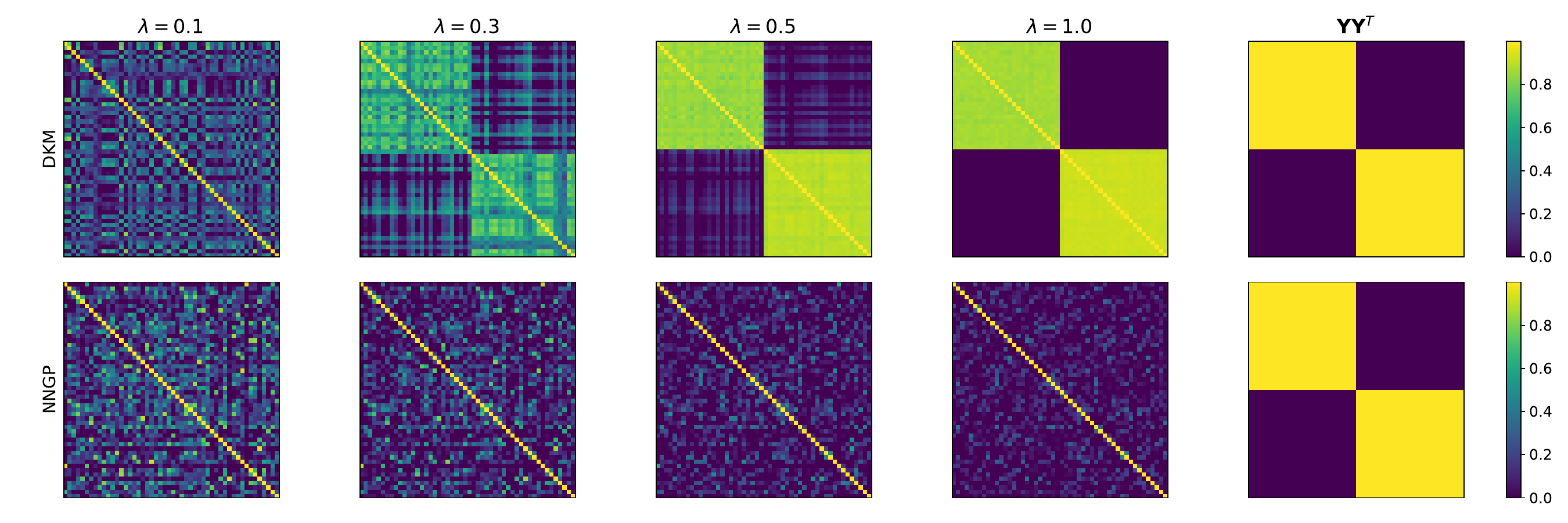}
    \caption{Final layer kernels for 2-layer linear graph convolutional DKMs (top row) and graph convolutional NNGPs (bottom row). The models are fit on a toy dataset with Gaussian random inputs and an~\erdosrenyi~adjacency (50 nodes, and edge probability 0.1). The label kernel is shown in the right column. The remaining columns show the effect of varying $\lambda$.}
    \label{fig:analyze_linear}
\end{figure*}

\begin{figure*}[t]
    \centering
    \includegraphics[width=\textwidth]{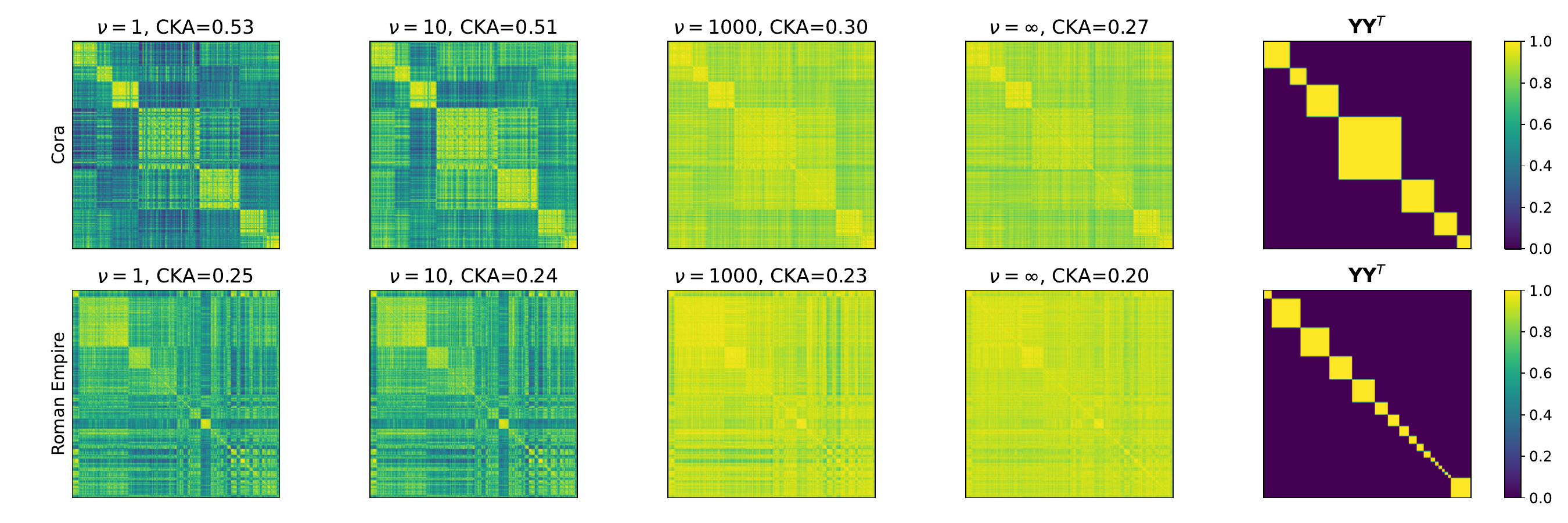}
    \caption{Final layer kernels~\cora~and~\romanempire~kernels for different regularizations \dof. The kernels are formed from 400 nodes sampled at random, and the `true' label kernel is shown on the rightmost column.}
    \label{fig:shaped}
\end{figure*}

\begin{figure*}[t]
    \centering
    \includegraphics[width=\textwidth]{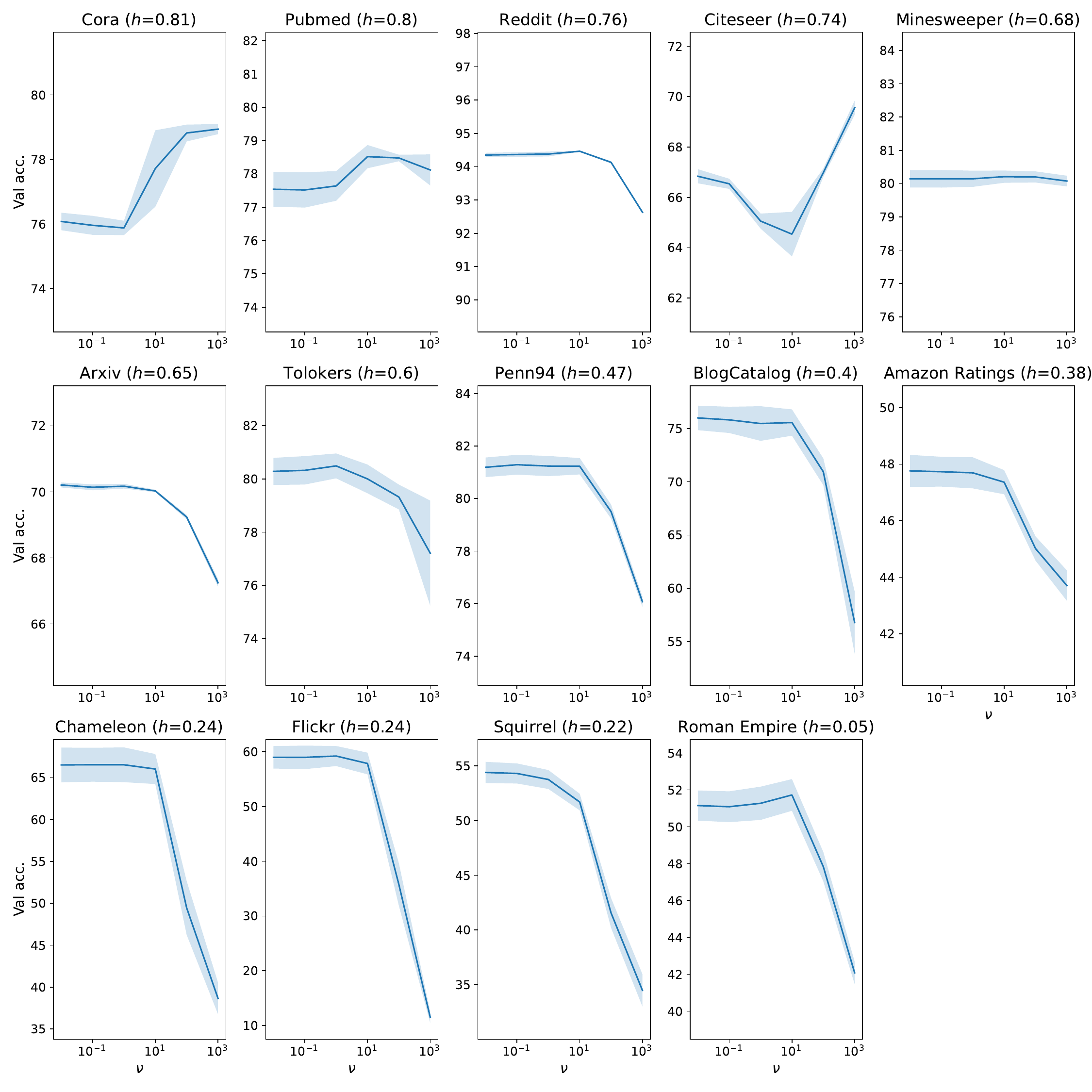}
    \caption{Validation accuracy at different regularization strengths ($\nu$) for each node classification dataset. The datasets are arranged by homophily ratio (denoted by $h$). The error bands denote $\pm 1$ standard deviation in the validation accuracy over several random seeds. For each subplot, we ensure the y-axis range is at least 8, so that the effect of $\nu$ is comparable between dataset.}
    \label{fig:val_acc_vs_dof_kipf}
\end{figure*}

Remarkably, it is possible to study graph convolutional DKMs analytically in the linear kernel case, which gives insight into representation learning in graphs.
Specifically, if we use a linear kernel ($\v K(\G) = \G$), set the regularization to 1 ($\nu_\ell=1$ for all $\ell$), we show in Appendix~\ref{app:closed_form_linear} that the Gram matrix representations optimizing the graph convolutional DKM objective at each layer $\ell$ are given by,
\begin{align}\label{eq:linear_gcdkm_solution}
\v G^\ell = \hA^{\ell-1}((\hA^{-L}\v G^{L+1}\hA^{-L})(\hA\v G^0\hA)^{-1})^{\ell/(L+1)} (\hA\v G^0 \hA)\hA^{\ell-1}.
\end{align}
Here $\v G^{0}$ and $\v G^{L+1}$ are the input and output kernels respectively (and are assumed to be known). We provide numerical verification of Eq.~\eqref{eq:linear_gcdkm_solution} by comparing it to a solution obtained via gradient descent in Figure~\ref{fig:linear_confirm_kernel}.

To ensure that the normalized adjacency is invertible, we compute Eq.~\eqref{eq:linear_gcdkm_solution} using $\hA_\lambda$ in place of $\hA$, which interpolates with the identity matrix via a parameter $\lambda\in[0,1]$,
\begin{align}\label{eq:res_adj}
\hA_\lambda = \lambda \v I +  (1-\lambda) \hA.
\end{align}
It is very natural to introduce this modified adjacency matrix in the graph setting, which incorporated by other prior work~\citep{luan2020complete}.
In particular, with $\lambda=0$, the graph mixup/convolution performs an equally weighted average over itself and all adjacent nodes.
This can be problematic.
For instance, consider a network consisting of two connected nodes.
After a single graph mixup, the two nodes will have the same features, which is unlikely to be desirable.
To fix this issue, we can use $\lambda>0$, which implies that the graph mixup is now a weighted average, with more weight on self-features.
This fixes the issue with our simple example network with two nodes: the resulting features are different even after a graph mixup.

We sought to compare the linear graph convolutional DKM kernel to the linear graph convolutional NNGP kernel,
\begin{align}\label{eq:linear_gcnnp_kernel}
\v G^\ell_{\text{GCNNGP}} = \hA^{\ell}{\v G}^0\hA^\ell.
\end{align}
While it is not practical to calculate Eq.~\eqref{eq:linear_gcdkm_solution} for large datasets (due to the inversion of a large adjacency matrix), a comparison is still possible with small datasets.
We compare the two kernels on a randomly generated an~\erdosrenyi~graph~\citep{erdos1963asymmetric} dataset of 50 nodes (with Gaussian features), and insert edges between nodes with probability 0.1. Half of the nodes are given a positive label and the remainder a negative label. We calculated the output layer kernels for $\lambda\in\{0.1,0.3,0.5,1\}$, and depict results in Figure~\ref{fig:analyze_linear}.
Note that the~\erdosrenyi~adjacency is not homophilous, because an edge is equally likely between both nodes with the same class and with a different class.

We see in Figure~\ref{fig:analyze_linear} (bottom row) that none of the graph convolutional NNGP kernels are aligned to the output kernel, $\v Y \v Y ^T$. This is because the graph convolutional NNGP kernel is a deterministic function of the inputs and adjacency, and is not learned based on the labels, $\v Y$. In contrast, the DKM is well-aligned with the output (Figure~\ref{fig:analyze_linear}, top row). Note we find that increasing $\lambda$ improves alignment of the graph convolutional DKM kernels to the output kernel. As $\lambda \rightarrow 0$, the adjacency matrix becomes singular, and it seems that this also involves a loss of representation learning, as the graph structure dominates the representation.
\section{Experiments}\label{sec:experiments}
In this section we investigate the effect of representation learning in graph networks by training graph convolutional DKMs on several datasets. We train on node classification datasets with varying homophily, as well as graph classification datasets. Our experiments show that representation learning helps to shape output kernels, and that datasets with lower homophily benefit more from representation learning than datasets with high homophily. We also compare the performance of graph convolutional DKMs to GCNs and graph convolutional NNGPs, and find that the extra flexibility afforded by the DKM framework closes the performance gap between graph convolutional NNGPs and GCNs.

In all experiments we use the ReLU/arccosine kernel non-linearity, and set the regularization $\nu_\ell=\nu$ such that it constant for all layers.
\subsection{Deep Kernel Machines Allow Shaping of Kernel Representations}\label{sec:shaping}
Similarly to the fully-connected DKM (\cite{dkm23}, Section 4.6), we demonstrate in Figure~\ref{fig:shaped} that the graph convolutional DKM shapes final layer kernel representations towards the label kernel, $\v Y\v Y^T$. We show alignment both qualitatively via the top-layer kernels, and quantitatively with the Centered Kernel Alignment (CKA)~\citep{cortes2012algorithms} metric. CKA is measured by,
\begin{align}
\mathrm{CKA}(\v K_1, \v K_2) = \frac{\Tr({\v K'_1  \v K'_2 })}{\sqrt{\Tr({\v K'_1 \v K'_1 })\Tr({\v K'_2 \v K'_2})}},
\end{align}
where $\v K'_{1/2}$ are centered version of the kernels $\v K_{1/2}$ to be compared. Higher CKA values indicate greater similarity between representations, with $\mathrm{CKA}(\v K_1,\v K_2)\in [0,1]$.

For this experiment, we trained graph convolutional DKMs for different values of $\nu\,\in\{10^0, 10^1, 10^3, \infty\}$ on~\cora~and~\romanempire~with the same random seed for 300 epochs. The models were trained using a simple 2-layer architecture with no residual connections or normalization/centering layers, and we used normalized adjacency $\hA_{\lambda=0.3}$ for the graph convolutions/mixups. The most suitable inducing point scheme was determined by a hyperparameter sweep over schemes for each dataset (see Section~\ref{sec:homo_and_rep_learn} and Appendix~\ref{app:experimental_details} for details). The kernels shown were obtained by taking the kernel representations of a randomly sampled 400 node subset of the last hidden layer, and the CKA similarity measure was calculated on the full final hidden layer kernel.

When~\dof~is 1 or 10, we clearly see block-diagonal structure mirroring $\v Y \v Y^T$, indicating ``kernel alignment''.
As~\dof~increases to 1000 or $\infty$, flexibility decreases and the kernel alignment appears to become weaker, and is confirmed by the CKA statistics.
Interestingly, this breakdown seems to involve all outputs becoming correlated, which is indicative of the rank-collapse-like phenomenon described by~\cite{oono2019graph}.
\subsection{Investigating the Relationship Between Homophily and Representation Learning}\label{sec:homo_and_rep_learn}
The graph convolutional NNGP kernel has a natural inductive bias for homophilous datasets; the graph convolutional NNGP pools representations from adjacent nodes, therefore in the extreme case that the connected components of the graph correspond to exactly the class labels, we expect its kernel to align with the labels. We expect the graph convolutional NNGP to be less effective for heterophilous datasets, because the graph structure is not informative for the labels. However, representation learning should help in the heterophilous case because we showed in Section~\ref{sec:shaping} that increasing flexibility helps shape the kernel representations.

We test the effect of representation learning empirically, training graph convolutional DKMs on several node classification benchmark datasets that exhibit varied levels of homophily. The dataset statistics can be found in Table~\ref{tab:node_classification_statistics}. We trained with different regularization strengths,~\dof~$\in\{0,10^{-2},10^{-1},10^0,$ $10^1,10^2,10^3\}$ to control the amount of representation learning. We also trained with two inducing point schemes (detailed in Section~\ref{sec:ind_point_schemes}), and for each dataset we selected the optimal scheme by calculating mean validation accuracy for each~\dof~and scheme. We used a 2-layer architecture, with the adjacency renormalization described by~\cite{kipf2016semi}, and no residual connections or normalization layers.

Figure~\ref{fig:val_acc_vs_dof_kipf} illustrates the empirical differences between NNGPs and DKMs, which we find to be highly dataset-dependent. For homophilous datasets (see top row of Figure~\ref{fig:val_acc_vs_dof_kipf}) like~\cora{}, performance tends to remain relatively stable or even improve with larger~\dof, suggesting that the standard architectural inductive biases are well-suited to these tasks. In contrast, heterophilous datasets such as~\romanempire~benefit greatly from smaller~\dof~(i.e. increased flexibility). This aligns with our theoretical understanding, as DKMs converge to NNGPs in the limit~\dof$\rightarrow \infty$. The lack of representation learning in NNGPs becomes particularly problematic for small $h$ (bottom row), leading to dramatic declines in accuracy of at least $\sim$10\% when comparing~\dof$=10^3$ to~\dof$=10^{-2}$. This suggests that for heterophilous tasks, where standard architectural inductive biases are less appropriate, the graph convolutional DKM's ability to learn representations allows it to compensate, while the graph convolution NNGP's inability to do so results in poor performance.
\begin{table*}[t]
  \centering
  \caption{Mean test accuracies (\%) $\pm$ one standard deviation on node classification datasets. Datasets are sorted by homophily, with most homophilous datasets at the top. We provide results for graph convolutional DKMs (GCDKM), graph convolutional NNGPs (GCNNGP) and GCNs. The best accuracy for each dataset is bolded. For datasets with multiple splits, the standard deviation is calculated using the means over splits. In some instances, the best graph convolutional DKM had $\nu=\infty$, which is denoted by $(*)$ (hence the sparse GCNNGP results are identical), and $(\dagger)$ denotes graph convolutional NNGP results from~\cite{niu2023graph}.}
  \vskip 0.15in
  \label{tab:final_perf}
\begin{tabular}{llllll}
\toprule
 & GCDKM & sparse GCNNGP & GCNNGP & GCN (no dropout) & GCN \\
\midrule
Cora & $81.1 \pm 0.2$ & $80.8 \pm 0.1$ & $\mathbf{82.8}$$^\dagger$ & $80.6 \pm 0.2$ & $81.1 \pm 0.3$ \\
Pubmed & $\mathbf{79.8 \pm 0.2}$$^*$ & $\mathbf{79.8 \pm 0.2}$ & $79.6$$^\dagger$ & $79.4 \pm 0.2$ & $79.5 \pm 0.1$ \\
Reddit & $\mathbf{96.2 \pm 0.0}$ & $93.5 \pm 0.0$ & $94.7 \pm 0.0$$^\dagger$ & $95.3 \pm 0.1$ & $95.8 \pm 0.1$ \\
Citeseer & $\mathbf{71.7 \pm 0.4}$$^*$ & $\mathbf{71.7 \pm 0.4}$ & $69.5$$^\dagger$ & $\mathbf{71.7 \pm 0.2}$ & $\mathbf{71.7 \pm 0.2}$ \\
Minesweeper & $\mathbf{85.9 \pm 0.4}$ & $84.3 \pm 0.3$ & --- & $85.6 \pm 0.4$ & $85.6 \pm 0.4$ \\
Arxiv & $\mathbf{70.8 \pm 0.2}$ & $68.9 \pm 0.2$ & $70.1 \pm 0.1$$^\dagger$ & $70.0 \pm 0.1$ & $70.6 \pm 0.2$ \\
Tolokers & $81.4 \pm 0.4$ & $79.0 \pm 0.1$ & --- & $81.6 \pm 0.7$ & $\mathbf{82.2 \pm 0.7}$ \\
Penn94 & $\mathbf{83.0 \pm 0.2}$ & $75.2 \pm 0.3$ & --- & $81.2 \pm 0.5$ & $81.5 \pm 1.8$ \\
BlogCatalog & $92.9 \pm 0.5$ & $92.8 \pm 0.5$ & --- & $93.6 \pm 0.6$ & $\mathbf{94.1 \pm 0.6}$ \\
Amazon Ratings & $49.2 \pm 0.5$ & $46.4 \pm 0.3$ & --- & $48.5 \pm 0.5$ & $\mathbf{49.8 \pm 0.6}$ \\
Flickr & $\mathbf{85.7 \pm 0.6}$ & $81.5 \pm 2.2$ & --- & $81.3 \pm 2.0$ & $82.6 \pm 0.7$ \\
Chameleon & $\mathbf{68.9 \pm 1.4}$ & $65.1 \pm 1.1$ & --- & $65.3 \pm 2.6$ & $65.3 \pm 2.6$ \\
Squirrel & $56.8 \pm 1.4$ & $38.6 \pm 1.2$ & --- & $57.3 \pm 1.4$ & $\mathbf{57.7 \pm 1.4}$ \\
Roman Empire & $79.2 \pm 0.4$ & $73.6 \pm 0.3$ & --- & $79.9 \pm 0.5$ & $\mathbf{82.8 \pm 0.5}$ \\
\bottomrule
\end{tabular}
\end{table*}

\begin{table*}[t]
  \centering
  \caption{Average test accuracies (\%) $\pm$ one standard deviation on graph classification datasets. Error bars are not statistically significant. GCN accuracies are sourced from~\cite{zhang2019hierarchical}($\star$) and~\cite{yang2020factorizable} ($\ddagger$).}
  \vskip 0.15in
  \label{tab:final_perf_graph}
\begin{tabular}{lllll}
\toprule
 & GCDKM & sparse GCNNGP & GCN \\
\midrule
Mutag & $86.6 \pm 3.8$ & $80.8 \pm 5.5$  & $85.6\pm 5.8^\ddagger$ \\
NCI1 & $75.3 \pm 1.0$ & $66.5 \pm 1.0$ & $76.3\pm 1.8^\star$ \\
NCI109 & $75.2 \pm 1.0$ & $65.9 \pm 1.3$ &  $75.9\pm1.8^\star$ \\
Proteins & $72.2 \pm 1.2$ & $67.6 \pm 1.4$ & $75.2\pm 3.6^\star$ \\
Mutagenicity & $79.1 \pm 0.7$ & $72.7 \pm 1.1$ & $79.8\pm 1.6^\star$\\
\bottomrule
\end{tabular}
\end{table*}

\subsection{Final Results}
To assess overall performance of the graph convolutional DKM, we optimized
hyperparameters for each dataset with a series of grid searches on a range of node and graph classification datasets. We kept the number of layers fixed at 2, and searched over inducing scheme, $\nu$, architecture, kernel centering, and finally the number of inducing points. The final test accuracies are shown in Table~\ref{tab:final_perf} and Table~\ref{tab:final_perf_graph}. We provide details of the datasets and the training procedure in the Appendix~\ref{app:train_details}.

In Table~\ref{tab:final_perf}, we see that the graph convolutional DKM performs similarly to the graph convolutional NNGP for homophilous node classification datasets, but less so for heterophilous datasets. We also see that the graph convolutional DKM performs similarly to the GCN for most datasets, closing the performance gap between infinite-width and finite-width networks. We attribute the instances where the GCN outperforms the graph convolutional DKM to dropout, as test accuracy of the graph convolutional DKM is similar to the GCN without dropout.
We also provide full-rank graph convolutional NNGP accuracies (where applicable) from~\cite{niu2023graph}. Their graph convolutional NNGP results are slightly different ours, most likely because ours is a sparse model with learned inducing points, whereas theirs is treated like a conventional GP.

In Table~\ref{tab:final_perf_graph}, we find that graph convolutional DKM performance is competitive with the GCN on graph classifications benchmarks, and uniformly better than the NNGP. The superior performance versus the NNGP is perhaps expected; the graph classification datasets are all molecule datasets, which tend to exhibit a mixture of homophily and heterophily~\citep{ye2022incorporating}, and thus the flexibility of the DKM ought to be very helpful.

Overall, we find that the graph convolutional DKM closes the gap between the NNGP and the GCN, supporting the hypothesis that representation learning is the key element lacking from the NNGP. This appears to be true regardless of task (e.g., node classification or graph classification). Further, our results demonstrate that the NNGP is well-suited to homophilous tasks, thus explaining why it is possible to obtain strong NNGP performance on some benchmark graph tasks~\citep{niu2023graph}, despite NNGPs generally being considered inferior GCNs.
\section{Limitations}
The key limitations of graph convolutional DKMs, like kernel methods in general, is scaling to large dataset sizes due to the cubic cost of naive methods. While we implemented an inducing point scheme to circumvent this limitation, it may be possible to improve on these methods.

Our current analysis considers standard graph convolutional architectures. While it is theoretically possible to develop DKM extensions by `kernelizing' alternative graph architectures~\citep{bo2021beyond,chien2020adaptive,velivckovic2017graph}, these extensions are non-trivial and beyond the scope of this work. We speculate that such extensions would improve performance across all methods (graph convolutional DKM/NNGP, and GCN) by incorporating additional inductive biases. The impact might be most pronounced for NNGPs, which lack representation learning capabilities and thus rely more heavily on built-in inductive biases. In contrast, graph convolutional DKMs and GCNs can compensate for imperfect inductive biases through representation learning.
\section{Conclusion and Discussion}
By leveraging the DKM framework, we have developed a flexible variant of an
infinite-width graph convolutional network. The graph convolutional DKM provides
a hyperparameter that allows us to interpolate between a flexible kernel machine
and the graph convolutional NNGP with fixed kernel. This feature of the DKM
enabled us to examine the importance of representation learning on graph node
classification tasks, and show competitive performance on graph classification
tasks. Remarkably, we found that some tasks benefit more from representation
learning than others. This finding helps explain the fact that NNGPs have
previously been shown to perform well on graph tasks, but not in other domains.
Moreover, the fact that our graph DKM model performs similarly to a GCN without
dropout suggests that the kind of flexibility introduced by the DKM is the
sensible kind, and may be used to advance kernel methods in the era of deep
learning.

Despite using an inducing point scheme, computational cost of the graph convolutional DKMs is
still relatively high. We expect that improving computational efficiency can
help improve performance, as it would allow for easier architecture exploration,
and training of bigger models. We also expect that more careful optimization
could boost performance even further since efficiency gains could enable larger models and longer training. However, we leave these issues for future work.
\newpage
\bibliography{main}

\begin{thebibliography}{58}
\providecommand{\natexlab}[1]{#1}
\providecommand{\url}[1]{\texttt{#1}}
\expandafter\ifx\csname urlstyle\endcsname\relax
  \providecommand{\doi}[1]{doi: #1}\else
  \providecommand{\doi}{doi: \begingroup \urlstyle{rm}\Url}\fi

\bibitem[Achten et~al.(2023)Achten, Tonin, Patrinos, and
  Suykens]{achten2023semi}
Sonny Achten, Francesco Tonin, Panagiotis Patrinos, and Johan~AK Suykens.
\newblock Semi-supervised classification with graph convolutional kernel
  machines.
\newblock \emph{arXiv preprint arXiv:2301.13764}, 2023.

\bibitem[Adlam et~al.(2023)Adlam, Lee, Padhy, Nado, and Snoek]{adlam2023kernel}
Ben Adlam, Jaehoon Lee, Shreyas Padhy, Zachary Nado, and Jasper Snoek.
\newblock Kernel regression with infinite-width neural networks on millions of
  examples.
\newblock \emph{arXiv preprint arXiv:2303.05420}, 2023.

\bibitem[Aitchison(2020)]{aitchison2020why}
Laurence Aitchison.
\newblock Why bigger is not always better: on finite and infinite neural
  networks.
\newblock In \emph{{ICML}}, 2020.

\bibitem[Aitchison et~al.(2021)Aitchison, Yang, and Ober]{aitchison2021deep}
Laurence Aitchison, Adam Yang, and Sebastian~W Ober.
\newblock Deep kernel processes.
\newblock In \emph{International Conference on Machine Learning}. PMLR, 2021.

\bibitem[Bengio et~al.(2013)Bengio, Courville, and
  Vincent]{bengio2013representation}
Yoshua Bengio, Aaron Courville, and Pascal Vincent.
\newblock Representation learning: A review and new perspectives.
\newblock \emph{IEEE transactions on pattern analysis and machine
  intelligence}, 35\penalty0 (8):\penalty0 1798--1828, 2013.

\bibitem[Bo et~al.(2021)Bo, Wang, Shi, and Shen]{bo2021beyond}
Deyu Bo, Xiao Wang, Chuan Shi, and Huawei Shen.
\newblock Beyond low-frequency information in graph convolutional networks.
\newblock In \emph{Proceedings of the AAAI conference on artificial
  intelligence}, volume~35, pp.\  3950--3957, 2021.

\bibitem[Bordelon \& Pehlevan(2023)Bordelon and Pehlevan]{bordelon2023self}
Blake Bordelon and Cengiz Pehlevan.
\newblock Self-consistent dynamical field theory of kernel evolution in wide
  neural networks.
\newblock \emph{Journal of Statistical Mechanics: Theory and Experiment},
  2023\penalty0 (11):\penalty0 114009, 2023.

\bibitem[Bronstein et~al.(2017)Bronstein, Bruna, LeCun, Szlam, and
  Vandergheynst]{bronstein2017geometric}
Michael~M Bronstein, Joan Bruna, Yann LeCun, Arthur Szlam, and Pierre
  Vandergheynst.
\newblock Geometric deep learning: going beyond euclidean data.
\newblock \emph{IEEE Signal Processing Magazine}, 34\penalty0 (4):\penalty0
  18--42, 2017.

\bibitem[Chien et~al.(2020)Chien, Peng, Li, and Milenkovic]{chien2020adaptive}
Eli Chien, Jianhao Peng, Pan Li, and Olgica Milenkovic.
\newblock Adaptive universal generalized pagerank graph neural network.
\newblock \emph{arXiv preprint arXiv:2006.07988}, 2020.

\bibitem[Cho \& Saul(2009)Cho and Saul]{cho2009kernel}
Youngmin Cho and Lawrence Saul.
\newblock Kernel methods for deep learning.
\newblock \emph{Advances in neural information processing systems}, 2009.

\bibitem[Cortes et~al.(2012)Cortes, Mohri, and
  Rostamizadeh]{cortes2012algorithms}
Corinna Cortes, Mehryar Mohri, and Afshin Rostamizadeh.
\newblock Algorithms for learning kernels based on centered alignment.
\newblock \emph{The Journal of Machine Learning Research}, 13:\penalty0
  795--828, 2012.

\bibitem[Cosmo et~al.(2021)Cosmo, Minello, Bronstein, Rodol{\`a}, Rossi, and
  Torsello]{cosmo2021graph}
Luca Cosmo, Giorgia Minello, Michael Bronstein, Emanuele Rodol{\`a}, Luca
  Rossi, and Andrea Torsello.
\newblock Graph kernel neural networks.
\newblock \emph{arXiv preprint arXiv:2112.07436}, 2021.

\bibitem[Damianou \& Lawrence(2013)Damianou and Lawrence]{damianou2013deep}
Andreas Damianou and Neil Lawrence.
\newblock Deep gaussian processes.
\newblock In \emph{Artificial Intelligence and Statistics}, 2013.

\bibitem[Erdos \& R{\'e}nyi(1963)Erdos and R{\'e}nyi]{erdos1963asymmetric}
Paul Erdos and Alfr{\'e}d R{\'e}nyi.
\newblock Asymmetric graphs.
\newblock \emph{Acta Math. Acad. Sci. Hungar}, 14\penalty0 (295-315):\penalty0
  3, 1963.

\bibitem[Fey \& Lenssen(2019)Fey and Lenssen]{Fey/Lenssen/2019}
Matthias Fey and Jan~E. Lenssen.
\newblock Fast graph representation learning with {PyTorch Geometric}.
\newblock In \emph{ICLR Workshop on Representation Learning on Graphs and
  Manifolds}, 2019.

\bibitem[Garriga-Alonso et~al.(2018)Garriga-Alonso, Rasmussen, and
  Aitchison]{garriga2018deep}
Adri{\`a} Garriga-Alonso, Carl~Edward Rasmussen, and Laurence Aitchison.
\newblock Deep convolutional networks as shallow gaussian processes.
\newblock \emph{arXiv preprint arXiv:1808.05587}, 2018.

\bibitem[Gupta \& Nagar(2018)Gupta and Nagar]{gupta2018matrix}
Arjun~K Gupta and Daya~K Nagar.
\newblock \emph{Matrix variate distributions}.
\newblock Chapman and Hall/CRC, 2018.

\bibitem[Hamilton et~al.(2017)Hamilton, Ying, and
  Leskovec]{hamilton2017inductive}
Will Hamilton, Zhitao Ying, and Jure Leskovec.
\newblock Inductive representation learning on large graphs.
\newblock \emph{Advances in neural information processing systems}, 2017.

\bibitem[Hu et~al.(2020{\natexlab{a}})Hu, Shen, Yang, and
  Shao]{hu2020infinitely}
Jilin Hu, Jianbing Shen, Bin Yang, and Ling Shao.
\newblock Infinitely wide graph convolutional networks: semi-supervised
  learning via gaussian processes.
\newblock \emph{arXiv preprint arXiv:2002.12168}, 2020{\natexlab{a}}.

\bibitem[Hu et~al.(2020{\natexlab{b}})Hu, Fey, Zitnik, Dong, Ren, Liu, Catasta,
  and Leskovec]{hu2020open}
Weihua Hu, Matthias Fey, Marinka Zitnik, Yuxiao Dong, Hongyu Ren, Bowen Liu,
  Michele Catasta, and Jure Leskovec.
\newblock Open graph benchmark: Datasets for machine learning on graphs.
\newblock \emph{Advances in neural information processing systems},
  33:\penalty0 22118--22133, 2020{\natexlab{b}}.

\bibitem[Jacot et~al.(2018)Jacot, Hongler, and Gabriel]{jacot2018neural}
Arthur Jacot, Cl{\'{e}}ment Hongler, and Franck Gabriel.
\newblock Neural tangent kernel: Convergence and generalization in neural
  networks.
\newblock In \emph{NeurIPS}, pp.\  8580--8589, 2018.

\bibitem[Kashima et~al.(2003)Kashima, Tsuda, and
  Inokuchi]{kashima2003marginalized}
Hisashi Kashima, Koji Tsuda, and Akihiro Inokuchi.
\newblock Marginalized kernels between labeled graphs.
\newblock In \emph{Proceedings of the 20th international conference on machine
  learning (ICML-03)}, pp.\  321--328, 2003.

\bibitem[Kipf \& Welling(2017)Kipf and Welling]{kipf2016semi}
Thomas~N Kipf and Max Welling.
\newblock Semi-supervised classification with graph convolutional networks.
\newblock \emph{International Conference on Learning Representations}, 2017.

\bibitem[LeCun et~al.(2015)LeCun, Bengio, and Hinton]{lecun2015deep}
Yann LeCun, Yoshua Bengio, and Geoffrey Hinton.
\newblock Deep learning.
\newblock \emph{nature}, 521\penalty0 (7553):\penalty0 436--444, 2015.

\bibitem[Lee et~al.(2017)Lee, Bahri, Novak, Schoenholz, Pennington, and
  Sohl-Dickstein]{lee2017deep}
Jaehoon Lee, Yasaman Bahri, Roman Novak, Samuel~S Schoenholz, Jeffrey
  Pennington, and Jascha Sohl-Dickstein.
\newblock Deep neural networks as gaussian processes.
\newblock \emph{arXiv preprint arXiv:1711.00165}, 2017.

\bibitem[Lee et~al.(2018)Lee, Bahri, Novak, Schoenholz, Pennington, and
  Sohl-Dickstein]{lee2018deep}
Jaehoon Lee, Yasaman Bahri, Roman Novak, Samuel~S Schoenholz, Jeffrey
  Pennington, and Jascha Sohl-Dickstein.
\newblock Deep neural networks as gaussian processes.
\newblock \emph{International Conference on Learning Representations}, 2018.

\bibitem[Lee et~al.(2020)Lee, Schoenholz, Pennington, Adlam, Xiao, Novak, and
  Sohl-Dickstein]{lee2020finite}
Jaehoon Lee, Samuel Schoenholz, Jeffrey Pennington, Ben Adlam, Lechao Xiao,
  Roman Novak, and Jascha Sohl-Dickstein.
\newblock Finite versus infinite neural networks: an empirical study.
\newblock \emph{Advances in Neural Information Processing Systems},
  33:\penalty0 15156--15172, 2020.

\bibitem[Lim et~al.(2021)Lim, Hohne, Li, Huang, Gupta, Bhalerao, and
  Lim]{lim2021large}
Derek Lim, Felix Hohne, Xiuyu Li, Sijia~Linda Huang, Vaishnavi Gupta, Omkar
  Bhalerao, and Ser~Nam Lim.
\newblock Large scale learning on non-homophilous graphs: New benchmarks and
  strong simple methods.
\newblock \emph{Advances in neural information processing systems},
  34:\penalty0 20887--20902, 2021.

\bibitem[Luan et~al.(2020)Luan, Zhao, Hua, Chang, and Precup]{luan2020complete}
Sitao Luan, Mingde Zhao, Chenqing Hua, Xiao-Wen Chang, and Doina Precup.
\newblock Complete the missing half: Augmenting aggregation filtering with
  diversification for graph convolutional networks.
\newblock \emph{arXiv preprint arXiv:2008.08844}, 2020.

\bibitem[Luan et~al.(2024)Luan, Hua, Xu, Lu, Zhu, Chang, Fu, Leskovec, and
  Precup]{luan2024graph}
Sitao Luan, Chenqing Hua, Minkai Xu, Qincheng Lu, Jiaqi Zhu, Xiao-Wen Chang,
  Jie Fu, Jure Leskovec, and Doina Precup.
\newblock When do graph neural networks help with node classification?
  investigating the homophily principle on node distinguishability.
\newblock \emph{Advances in Neural Information Processing Systems}, 36, 2024.

\bibitem[MacKay et~al.(1998)]{mackay1998introduction}
David~JC MacKay et~al.
\newblock Introduction to gaussian processes.
\newblock \emph{NATO ASI series F computer and systems sciences}, 168:\penalty0
  133--166, 1998.

\bibitem[Matthews et~al.(2018)Matthews, Rowland, Hron, Turner, and
  Ghahramani]{matthews2018gaussian}
Alexander G de~G Matthews, Mark Rowland, Jiri Hron, Richard~E Turner, and
  Zoubin Ghahramani.
\newblock Gaussian process behaviour in wide deep neural networks.
\newblock \emph{arXiv preprint arXiv:1804.11271}, 2018.

\bibitem[Maurya et~al.(2021)Maurya, Liu, and Murata]{maurya2021improving}
Sunil~Kumar Maurya, Xin Liu, and Tsuyoshi Murata.
\newblock Improving graph neural networks with simple architecture design.
\newblock \emph{arXiv preprint arXiv:2105.07634}, 2021.

\bibitem[Milsom et~al.(2023)Milsom, Anson, and Aitchison]{milsom2023cdkm}
Edward Milsom, Ben Anson, and Laurence Aitchison.
\newblock Convolutional deep kernel machines, 2023.

\bibitem[Neal(1996)]{neal1996bayesian}
R.~M. Neal.
\newblock \emph{Bayesian Learning for Neural Networks, Vol. 118 of Lecture
  Notes in Statistics}.
\newblock Springer-Verlag, 1996.

\bibitem[Niu et~al.(2023)Niu, Anitescu, and Chen]{niu2023graph}
Zehao Niu, Mihai Anitescu, and Jie Chen.
\newblock Graph neural network-inspired kernels for gaussian processes in
  semi-supervised learning.
\newblock \emph{arXiv preprint arXiv:2302.05828}, 2023.
\newblock In press.

\bibitem[Novak et~al.(2018)Novak, Xiao, Lee, Bahri, Yang, Hron, Abolafia,
  Pennington, and Sohl-Dickstein]{novak2018bayesian}
Roman Novak, Lechao Xiao, Jaehoon Lee, Yasaman Bahri, Greg Yang, Jiri Hron,
  Daniel~A Abolafia, Jeffrey Pennington, and Jascha Sohl-Dickstein.
\newblock Bayesian deep convolutional networks with many channels are gaussian
  processes.
\newblock \emph{arXiv preprint arXiv:1810.05148}, 2018.

\bibitem[Oono \& Suzuki(2019)Oono and Suzuki]{oono2019graph}
Kenta Oono and Taiji Suzuki.
\newblock Graph neural networks exponentially lose expressive power for node
  classification.
\newblock \emph{arXiv preprint arXiv:1905.10947}, 2019.

\bibitem[Platonov et~al.(2023)Platonov, Kuznedelev, Diskin, Babenko, and
  Prokhorenkova]{platonov2023critical}
Oleg Platonov, Denis Kuznedelev, Michael Diskin, Artem Babenko, and Liudmila
  Prokhorenkova.
\newblock A critical look at the evaluation of gnns under heterophily: are we
  really making progress?
\newblock \emph{arXiv preprint arXiv:2302.11640}, 2023.

\bibitem[Pleiss \& Cunningham(2021)Pleiss and
  Cunningham]{pleiss2021limitations}
Geoff Pleiss and John~P Cunningham.
\newblock The limitations of large width in neural networks: A deep gaussian
  process perspective.
\newblock \emph{Advances in Neural Information Processing Systems},
  34:\penalty0 3349--3363, 2021.

\bibitem[Salimbeni \& Deisenroth(2017)Salimbeni and
  Deisenroth]{salimbeni2017doubly}
Hugh Salimbeni and Marc Deisenroth.
\newblock Doubly stochastic variational inference for deep gaussian processes.
\newblock \emph{Advances in neural information processing systems}, 2017.

\bibitem[Scarselli et~al.(2008)Scarselli, Gori, Tsoi, Hagenbuchner, and
  Monfardini]{scarselli2008graph}
Franco Scarselli, Marco Gori, Ah~Chung Tsoi, Markus Hagenbuchner, and Gabriele
  Monfardini.
\newblock The graph neural network model.
\newblock \emph{IEEE transactions on neural networks}, 20\penalty0
  (1):\penalty0 61--80, 2008.

\bibitem[Shankar et~al.(2020)Shankar, Fang, Guo, Fridovich-Keil, Ragan-Kelley,
  Schmidt, and Recht]{shankar2020neural}
Vaishaal Shankar, Alex Fang, Wenshuo Guo, Sara Fridovich-Keil, Jonathan
  Ragan-Kelley, Ludwig Schmidt, and Benjamin Recht.
\newblock Neural kernels without tangents.
\newblock In \emph{International conference on machine learning}, pp.\
  8614--8623. PMLR, 2020.

\bibitem[Shervashidze \& Borgwardt(2009)Shervashidze and
  Borgwardt]{shervashidze2009fast}
Nino Shervashidze and Karsten Borgwardt.
\newblock Fast subtree kernels on graphs.
\newblock \emph{Advances in neural information processing systems}, 22, 2009.

\bibitem[Shervashidze et~al.(2009)Shervashidze, Vishwanathan, Petri, Mehlhorn,
  and Borgwardt]{shervashidze2009efficient}
Nino Shervashidze, SVN Vishwanathan, Tobias Petri, Kurt Mehlhorn, and Karsten
  Borgwardt.
\newblock Efficient graphlet kernels for large graph comparison.
\newblock In \emph{Artificial intelligence and statistics}, pp.\  488--495.
  PMLR, 2009.

\bibitem[Veli{\v{c}}kovi{\'c} et~al.(2017)Veli{\v{c}}kovi{\'c}, Cucurull,
  Casanova, Romero, Lio, and Bengio]{velivckovic2017graph}
Petar Veli{\v{c}}kovi{\'c}, Guillem Cucurull, Arantxa Casanova, Adriana Romero,
  Pietro Lio, and Yoshua Bengio.
\newblock Graph attention networks.
\newblock \emph{arXiv preprint arXiv:1710.10903}, 2017.

\bibitem[Vyas et~al.(2023)Vyas, Atanasov, Bordelon, Morwani, Sainathan, and
  Pehlevan]{vyas2023feature}
Nikhil Vyas, Alexander Atanasov, Blake Bordelon, Depen Morwani, Sabarish
  Sainathan, and Cengiz Pehlevan.
\newblock Feature-learning networks are consistent across widths at realistic
  scales.
\newblock \emph{arXiv preprint arXiv:2305.18411}, 2023.

\bibitem[Walker \& Glocker(2019)Walker and Glocker]{walker2019graph}
Ian Walker and Ben Glocker.
\newblock Graph convolutional gaussian processes.
\newblock In \emph{International Conference on Machine Learning}, pp.\
  6495--6504. PMLR, 2019.

\bibitem[Williams(1996)]{williams1996computing}
Christopher Williams.
\newblock Computing with infinite networks.
\newblock In M.C. Mozer, M.~Jordan, and T.~Petsche (eds.), \emph{Advances in
  Neural Information Processing Systems}, volume~9. MIT Press, 1996.
\newblock URL
  \url{https://proceedings.neurips.cc/paper_files/paper/1996/file/ae5e3ce40e0404a45ecacaaf05e5f735-Paper.pdf}.

\bibitem[Yanardag \& Vishwanathan(2015)Yanardag and
  Vishwanathan]{yanardag2015deep}
Pinar Yanardag and SVN Vishwanathan.
\newblock Deep graph kernels.
\newblock In \emph{Proceedings of the 21th ACM SIGKDD international conference
  on knowledge discovery and data mining}, 2015.

\bibitem[Yang et~al.(2023)Yang, Robeyns, Milsom, Schoots, Anson, and
  Aitchison]{dkm23}
Adam~X. Yang, Maxime Robeyns, Edward Milsom, Nandi Schoots, Ben Anson, and
  Laurence Aitchison.
\newblock A theory of representation learning in deep neural networks gives a
  deep generalisation of kernel methods.
\newblock \emph{International Conference on Machine Learning}, 2023.
\newblock In press.

\bibitem[Yang(2019)]{yang2019scaling}
Greg Yang.
\newblock Scaling limits of wide neural networks with weight sharing: Gaussian
  process behavior, gradient independence, and neural tangent kernel
  derivation.
\newblock \emph{arXiv preprint arXiv:1902.04760}, 2019.

\bibitem[Yang \& Hu(2021)Yang and Hu]{yang2021tensor}
Greg Yang and Edward~J. Hu.
\newblock Tensor programs {IV:} feature learning in infinite-width neural
  networks.
\newblock In \emph{{ICML}}, volume 139 of \emph{Proceedings of Machine Learning
  Research}, pp.\  11727--11737. {PMLR}, 2021.

\bibitem[Yang et~al.(2020)Yang, Feng, Song, and Wang]{yang2020factorizable}
Yiding Yang, Zunlei Feng, Mingli Song, and Xinchao Wang.
\newblock Factorizable graph convolutional networks.
\newblock \emph{Advances in Neural Information Processing Systems},
  33:\penalty0 20286--20296, 2020.

\bibitem[Ye et~al.(2022)Ye, Yang, Medya, and Singh]{ye2022incorporating}
Wei Ye, Jiayi Yang, Sourav Medya, and Ambuj Singh.
\newblock Incorporating heterophily into graph neural networks for graph
  classification.
\newblock \emph{arXiv preprint arXiv:2203.07678}, 2022.

\bibitem[Zhang et~al.(2019)Zhang, Bu, Ester, Zhang, Yao, Yu, and
  Wang]{zhang2019hierarchical}
Zhen Zhang, Jiajun Bu, Martin Ester, Jianfeng Zhang, Chengwei Yao, Zhi Yu, and
  Can Wang.
\newblock Hierarchical graph pooling with structure learning.
\newblock \emph{arXiv preprint arXiv:1911.05954}, 2019.

\bibitem[Zhu et~al.(2020)Zhu, Yan, Zhao, Heimann, Akoglu, and
  Koutra]{zhu2020beyond}
Jiong Zhu, Yujun Yan, Lingxiao Zhao, Mark Heimann, Leman Akoglu, and Danai
  Koutra.
\newblock Beyond homophily in graph neural networks: Current limitations and
  effective designs.
\newblock \emph{Advances in neural information processing systems},
  33:\penalty0 7793--7804, 2020.

\bibitem[Zhu et~al.(2021)Zhu, Rossi, Rao, Mai, Lipka, Ahmed, and
  Koutra]{zhu2021graph}
Jiong Zhu, Ryan~A Rossi, Anup Rao, Tung Mai, Nedim Lipka, Nesreen~K Ahmed, and
  Danai Koutra.
\newblock Graph neural networks with heterophily.
\newblock In \emph{Proceedings of the AAAI conference on artificial
  intelligence}, volume~35, pp.\  11168--11176, 2021.

\end{thebibliography}
\bibliographystyle{tmlr}

\newpage
\appendix
\section{Full Rank Graph Convolutional Deep Kernel Machines}\label{app:gcdkm_deriv}
Here, we outline a derivation for the graph convolutional DKM objective. We start with a DGP with graph mixups,
\begin{align}\label{eq:graph_dgp}
\P(\v F^\ell\mid \v F^{\ell-1}) &= \prod_{\lambda=1}^{N_\ell} \normal(\v f^\ell_\lambda; \v 0, \Ahat \v K_\text{features}(\v F^{\ell-1}) \Ahat^T),
\end{align}
where $\v f^\ell_\lambda$ are columns of features,
\begin{align}
\v F^\ell = (\v f^\ell_1 \cdots \v f^\ell_{N_\ell}).
\end{align}
We take $P$ to be the number of datapoints, and $N_\ell$ the number of features at each layer $\ell$, so that $\v F^\ell \in\mathbb{R}^{P\times N_\ell}$.
The log-posterior over Gram matrices $\v G^\ell = \tfrac{1}{N_\ell}\v F^\ell (\v F^\ell)^T$ is equivalent to,
\begin{align}\label{eq:gdgp_posterior}
\log\P(\v G^1,\ldots,\v G^L\mid \v X,\v Y) = \log\P(\v Y \mid \G^L) + \sum_{\ell=1}^L \log\P(\v G^{\ell}\mid \v G^{\ell-1}),
\end{align}
with $\v G^0$ the being a kernel formed from the inputs $\v X$.
Following Section~\ref{sec:dgps_in_terms_of_grams}, we assume that $\v K_\text{features}(\cdot)$ is a kernel function that can be calculated in terms of Gram matrices, i.e.,  $\exists \v K$ such that $\v K(\G^\ell) = \v K_\text{features}(\v F^\ell)$. Examples of suitable kernels are the arccosine kernel~\citep{cho2009kernel} or the squared exponential kernel.

Since $\v F^\ell$ is multivariate Gaussian, $\G^\ell$ is Wishart distributed by its definition, and we can write down its log density,
\begin{align}
\log\P(\v G^{\ell}\mid &\v G^{\ell-1}) = \tfrac{N_\ell - P - 1}{2}\log\abs{\v G^\ell} \nonumber\\
&-\tfrac{N_\ell}{2}\log\abs{\hA\v K(\v G^{\ell-1})\hA^T}  - \tfrac{N_\ell}{2}\Tr\{{(\hA\v K(\v G^{\ell-1})\hA^T)^{-1}\v G^\ell}\} + \alpha_\ell.
\end{align}
The constant $\alpha_\ell$ is the normalizing constant, and we assume that $\hA$ is full rank.
\cite{dkm23} showed that $\alpha_\ell$ satisfies the following scaling law as we increase width: $\lim_{N_\ell\rightarrow\infty}\alpha_\ell/N_\ell = \mathrm{const}$. Therefore if $N_\ell = N \nu_\ell$, it follows that,
\begin{align}\label{eq:dgdp_limit_gram}
\lim_{N\rightarrow\infty}\tfrac{1}{N}\log\P(\v G^{\ell}\mid &\v G^{\ell-1}) = \mathrm{const} + \nu_\ell\KL\left(\normal(\v 0,\v G^\ell)\mid\mid \normal(\v 0, \hA\v K(\v G^{\ell-1})\hA^T)\right).
\end{align}
The Bayesian representation learning limit~\citep{dkm23} dictates that we scale the final layer alongside the hidden layers (this is unlike the usual NNGP limit~\citep{lee2017deep} which only scales the intermediate layers). To do so, we duplicate the outputs so that there are $N$ copies of the labels, i.e. $\v{\tilde{Y}} = (\v Y \cdots \v Y)$. Since channels are assumed to be i.i.d., we have,
\begin{align}\label{eq:dgdp_ytilde}
\P(\v{\tilde Y} \mid \G^L ) = \P(\v Y \mid \G^L)^N.
\end{align}
We now consider the posterior of this network with wide hidden layers, wide outputs, and labels $\v{\tilde Y}$. Combining Eqs.~\eqref{eq:gdgp_posterior},~\eqref{eq:dgdp_limit_gram}, and~\eqref{eq:dgdp_ytilde}, and letting $N_\ell = N\nu_\ell$, we have that the scaled log-posterior over Gram matrices satisfies,
\begin{align}\label{eq:gcdkm_obj_fr}
\lim_{N\rightarrow\infty}&\tfrac{1}{N}\log\P(\v G^1,\ldots,\v G^L\mid \v X,\v{\tilde Y})\nonumber\\
&=\log\P(\v Y \mid \G^L)\nonumber
+\sum_{\ell=1}^L \nu_\ell\KL\left(\normal(\v 0,\v G^\ell)\mid\mid \normal(\v 0, \hA\v K(\v G^{\ell-1})\hA^T)\right) + \mathrm{const} \\
&:=\mathcal{L}(\G^1,\ldots,\G^L).
\end{align}
Finally, assuming that the features are sufficiently well-behaved, we note that the Gram matrices are deterministic in the limit $N\rightarrow\infty$, since they are a sum of i.i.d. terms,
\begin{align}\label{eq:deterministic_grams}
\G^\ell &= \tfrac{1}{N_\ell} \sum_{\lambda=1}^{N_\ell}\v f^\ell(\v f^\ell)^T
\implies \lim_{N\rightarrow\infty} \G^\ell = \mathrm{const}.
\end{align}
The consequence of Eq.~\eqref{eq:deterministic_grams} is that the limiting posterior of the Gram matrices is a point distribution. But we know that the Gram matrices on which this point distribution is centered on should maximize the posterior, and hence the scaled log-posterior. Therefore, we seek to maximize Eq.~\eqref{eq:gcdkm_obj_fr}, the scaled log-posterior. This is exactly the graph convolutional DKM objective we declared in Section~\ref{sec:gcdkms}, up to an additive constant.
\section{Inducing Point Approximations}\label{app:sec:inducing_point_appendix}
The graph convolutional DKM objective is difficult to optimize directly due to quadratic memory costs and cubic time complexity. To avoid these large costs, we use an inducing point approximation similar to the one described in~\citep{dkm23}. We review in this Appendix how to construct an inducing point objective in a fully-connect setting, and then extend to the graph setting.

\subsection{An Inducing Point Objective}
Sparse DKMs as derived by~\cite{dkm23} are inspired by the sparse DGP literature~\citep{damianou2013deep,salimbeni2017doubly}. In essence, we place an approximate posterior on a set of sparse inducing-points/features and derive an objective to optimize these inducing points by finding a lower bound on the evidence with variational inference (VI). In this subsection we describe the procedure.

We introducing some inducing-points/features $\Fi^\ell\in \reals^{\ppi\times N_{\ell}}$ in addition to the `real' test/training features $\Ft^\ell\in\reals^{\ppt \times N_\ell}$ at each layer in the following prior model,
\begin{subequations}\label{eq:dgp_bayesian_toplayer}
\begin{align}
\v F^\ell :&= \begin{pmatrix}
\Fi^\ell \\ \Ft^\ell
\end{pmatrix},\\
\pr(\v F^\ell \mid \v F^{\ell -1}) &= \prod_{\lambda=1}^{N_{\ell}}\normal(\v f_\lambda^\ell; \v 0, \v K_\text{features}(\v F^{\ell -1})),\\
\pr(\v W) &= \prod_{\lambda=1}^{\nu_{L+1}}\normal(\v w_\lambda; \v 0, \v I),\\
\pr(\v Y \mid \v K^L, \v W) &= \prod_{i=1}^{\ppt}\text{Categorical}(\v y_i; \mathrm{softmax}(\v K^L_\mathrm{ti} \mathrm{chol}(\v K^L_\mathrm{ii})^{-1}\v W)_i),
\end{align}
\end{subequations}
where $\v K^L$ is the kernel applied to the final hidden layer features $\v F^L$, with $\Kii^L$ being the inducing/inducing sub-block and $\Kti^L$ being the training/test-inducing sub-block. Eq.~\eqref{eq:dgp_bayesian_toplayer} is similar to the DGP in Eq.~\eqref{eq:dgp}, but it has a different top-layer. The motivation for using the top-layer in Eq.~\eqref{eq:dgp_bayesian_toplayer} is efficiency; it is more efficient than a categorical GP under the parameterization described in Appendix~\ref{app:parameterization}.

To perform variational inference, we place a Gaussian approximate posterior on the inducing hidden-layer features,
\begin{align}
\Q(\Fi^\ell) = \prod_{\lambda=1}^{N_\ell}\normal((\v f_\mathrm{i}^\ell)_\lambda; \v 0, \Gii),
\end{align}
and the following approximate posterior on the top-layer,
\begin{align}\label{eq:top_layer_approx_posterior}
\Q(\v W) = \prod_{\lambda=1}^{\nu_{L+1}}\normal(\v w_\lambda; \v \mu_\lambda, \v \Sigma).
\end{align}
Thus, $\Gii^1,\ldots, \Gii^L$, $\v \mu_1,\ldots,\v \mu_{\nu_{L+1}}$, and $\v \Sigma$ are the variational parameters.
We can write down the distribution of the test/training features conditional on the inducing features using standard multivariate Gaussian rules,
\begin{align}\label{eq:ft_cond_fi}
\P(\Ft^\ell\mid \Fi^\ell, \v F^{\ell-1}) &= \normal\left(\Ft^\ell; \Kti\Kii^{-1}\Fi^\ell, \Ktt - \Kti\Kii^{-1}\Kit\right),
\end{align}
where by $\Kii$, $\Kti$, and $\Ktt$ we refer to inducing-inducing, test/train-inducing, and test/train-test/train blocks respectively of $\v K(\v F^{\ell-1})$.
We use Eq.~\eqref{eq:ft_cond_fi} to construct an approximate posterior over the test/training features.
The prior at each layer is equivalent to,
\begin{align}
\P(\v F^\ell\mid \v F^{\ell-1}) = \P(\Ft^\ell\mid \Fi^\ell, \v F^{\ell-1})\P(\Fi^\ell\mid \Fi^{\ell-1}),
\end{align}
and we select the full approximate posterior at each hidden layer to be
\begin{align}
\Q(\v F^\ell\mid \v F^{\ell-1}) = \P(\Ft^\ell\mid \Fi^\ell, \v F^{\ell-1})\Q(\Fi^\ell).
\end{align}
This choice of approximate posterior is a classic one; the common term, $\P(\Ft^\ell\mid \Fi^\ell, \v F^{\ell-1})$, is going to lead to cancellation in the ELBO.
At the top layer, we use an i.i.d. Gaussian approximate posterior for the weights $\v W$,
\begin{align}
\Q(\v W) = \prod_{\lambda=1}^{\nu_{L+1}}\normal(\v w_\lambda; \v \mu_\lambda, \v \Sigma).
\end{align}

Similarly to Eq.~\eqref{eq:dgdp_ytilde} in the full-rank derivation, we duplicate the entries in the top-layer $N$ times and make an i.i.d. assumption, such that  $\v{\tilde Y} = (\v Y \cdots \v Y)\in \mathbb{R}^{\ppt \times N \nu_{L+1}}$  and  $\v{\tilde W} = (\v W \cdots \v W)\in \mathbb{R}^{\ppi \times N \nu_{L+1}}$.
Note that both the prior and the approximate posterior factorize across layers, which allows us to simplify the ELBO,
\begin{subequations}
\begin{align}
\text{ELBO} &=\mathbb{E}_{\Q}\bigg[\log\frac{\P({\v{\tilde Y}}, \v X, \v F^1,\ldots,\v F^L, \v{\tilde W})}{\Q(\v F^1,\ldots,\v F^L, \v{\tilde W})}\bigg]\\
&=\mathbb{E}_{\Q}\bigg[\log\frac{\P(\v{\tilde Y}, \v{\tilde W} \mid \v F^L)}{\Q(\v{\tilde W})} + \sum_{\ell=1}^L\log\frac{\P(\v F^\ell \mid \v F^{\ell-1})}{\Q(\v F^\ell)}\bigg]\\
&= \mathbb{E}_{\Q}\bigg[\log\frac{\P(\v{\tilde Y}, \v{\tilde W}\mid \v K^L)}{\Q(\v{\tilde W})}\bigg] + \sum_{\ell=1}^L\mathbb{E}_{\Q}\bigg[\log\frac{\P(\Fi^\ell \mid \Fi^{\ell-1})}{\Q(\Fi^\ell)}\bigg].
\end{align}
\end{subequations}
In the limit $N\rightarrow\infty$, where $N_\ell = N \nu_\ell$, the covariances at each layer, $N_\ell^{-1}\Fi^\ell(\Fi^\ell)^T$, collapse to become equal to $\Gii^\ell$. It can be shown that the hidden-layer terms in the ELBO simplify,
\begin{align}
\frac{1}{N}\mathbb{E}_{\Q}\bigg[\log\frac{\P(\Fi^\ell \mid \Fi^{\ell-1})}{\Q(\Fi^\ell)}\bigg] \rightarrow -\nu_\ell \KL\left(\normal(\v 0, \v G_\text{ii}^\ell)\mid\mid\normal(\v 0, \v  K(\v G_\text{ii}^{\ell-1}))\right).
\end{align}
Additionally, $\forall N$ the term involving $\v{\tilde Y}$ satisfies,
\begin{subequations}
\begin{align}
\frac{1}{N}\mathbb{E}_{\Q}\bigg[\log\frac{\P(\v{\tilde Y}, \v{\tilde W} \mid \v K^{L})}{\Q(\v{\tilde W})}\bigg] &=\mathbb{E}_{\Q}\bigg[\log \P(\v Y \mid \v K^{L},\v W) + \log\frac{\P(\v W)}{\Q(\v W)}\bigg]\\
&=\mathbb{E}_{\Q}[\log \P(\v Y \mid \v K^{L}, \v W)] - \sum_{\lambda=1}^{\nu_{L+1}}\KL\left(\normal({\v \mu}_\lambda, \v \Sigma)\mid\mid\normal(\v 0, \v I)\right),
\end{align}
\end{subequations}
which can easily be estimated by sampling $\v W$, since in the limit the kernel $\v K^L = \v K(\v G^L)$ is constant.

We define the sparse DKM objective to be the limiting ELBO scaled by $N^{-1}$,
\begin{equation}\label{eq:sparse_dkm_obj}
\begin{split}
   \lim_{N\rightarrow\infty} \frac{1}{N}\text{ELBO}:=\mathcal L_\text{sparse} = &\;\mathbb{E}_{\Q}[\log\P(\v Y \mid \v K(\v G^L), \v W)] \\
   &-\sum_{\lambda=1}^{\nu_{L+1}}\KL\left(\normal({\v \mu}_\lambda, \v \Sigma)\mid\mid\normal(\v 0, \v I)\right)\\
   &-\sum_{\ell=1}^L\nu_\ell \KL\left(\normal(\v 0, \v G_\text{ii}^\ell)\mid\mid\normal(\v 0, \v  K(\v G_\text{ii}^{\ell-1}))\right).
\end{split}
\end{equation}
Notice that the likelihood term in Eq.~\eqref{eq:sparse_dkm_obj} requires the full kernel. We calculate this recursively. Eq.~\eqref{eq:ft_cond_fi} tells us that,
\begin{align}
\Ft^\ell = \Kti\Kii^{-1}\Fi^\ell + (\Ktt - \Kti \Kii^{-1} \Kit)^{1/2} \v \Xi,
\end{align}
where $\v \Xi$ is a matrix of i.i.d. standard Gaussian noise, and $\v K = \v K(\v G^{\ell-1}) = [\Kii\ \, \Kit;\, \Kti\,\ \Ktt]$. Therefore,
\begin{align}
\Gti^\ell &= \lim N_\ell^{-1}\Ft^\ell (\Fi^\ell)^T  = \Kti \Kii^{-1}\Gii^\ell,\\
\Gtt^\ell &= \lim N_\ell^{-1}\Ft^\ell (\Ft^\ell)^T  = \Ktt  -  \Kti \Kii^{-1}\Kit + \Kti \Kii^{-1}\Gii^\ell\Kii^{-1}\Kit.
\end{align}

\subsection{Sparse Graph Convolutional Deep Kernel Machines}\label{sec:graph_ind_point_scheme}

To use a graph convolutional DKM, we need to be able to compute the graph kernel ${\v G \mapsto \v A \v K(\v G)\v A^T}$ with our inducing points.
We already have adjacency information regarding the training/test nodes, but not the inducing nodes. Therefore we assume we have an adjacency between inducing points, $\Aii$, as well as adjacency information between inducing points and training/test points, $\Ait = \Ati^T$. For notational consistency, we denote the adjacency of the data as $\Att$.

The graph convolutional kernel over Gram matrices when using an inducing scheme becomes
\begin{align}\label{eq:sparsegcnngp}
{\mathbf K}_{\text{GC}}(\v G) =
\begin{pmatrix}
\Aii & \Ait\\
\Ati & \Att
\end{pmatrix}\v K(\v G) \begin{pmatrix}
\Aii & \Ait\\
\Ati & \Att
\end{pmatrix}^T.
\end{align}
where $\v K(\v G)$ is a ``base kernel'' (e.g. the arccosine kernel) applied to $\v G$. We derive this kernel for an graph convolutional NNGP concretely in Appendix~\ref{app:kernelderiv}. Substituting this graph kernel into the intermediate layers of the sparse DKM in Eq.~\eqref{eq:sparse_dkm_obj} gives us a sparse graph convolutional DKM objective,
\begin{equation}\label{eq:sparse_gcdkm_obj}
\begin{split}
   \mathcal L_\text{GC-sparse} = &\;\mathbb{E}_{\v W \sim \Q}[\log\P(\v Y \mid \v K(\v G^L), \v W)] \\
   &-\sum_{\lambda=1}^{\nu_{L+1}}\KL\left(\normal({\v \mu}_\lambda, \v \Sigma)\mid\mid\normal(\v 0, \v I)\right)\\
   &-\sum_{\ell=1}^L\nu_\ell \KL\left(\normal(\v 0, \v G_\text{ii}^\ell)\mid\mid\normal(\v 0, \Aii\v  K(\v G_\text{ii}^{\ell-1})\Aii^T)\right),
\end{split}
\end{equation}
Again, the Gram matrices are calculated recursively in the forward pass with,
\begin{align}
\Gti^\ell &=  \Kti \Kii^{-1}\Gii^\ell,\\
\Gtt^\ell &=  \Ktt  -  \Kti \Kii^{-1}\Kit + \Kti \Kii^{-1}\Gii^\ell\Kii^{-1}\Kit,
\end{align}
and we then apply the graph convolutional kernel $\v K_\text{GC}(\cdot)$ to $\v G^\ell = [\Gii^\ell\ \, \Git^\ell;\, \Gti^\ell\,\ \Gtt^\ell]$.

We considered two choices for $\Aii$ and $\Ati$/$\Ait$, namely
\begin{enumerate}[(1)]
\item (intra-domain) using adjacency information from inducing points sampled from the training set,
\item (inter-domain) treating inducing points as independent of other inducing points and the training/test points, i.e. $\Aii = \v I $, $\Ati = \v 0$.
\end{enumerate}
In the case where the dataset is a single graph, we use a minibatch size of 1, and the `minibatch' becomes all nodes in the graph. In the case where the dataset contains several graphs, the graphs can be batched together.
Naively, this would involve creating a single large adjacency matrix that describes adjacency information for graphs in the minibatch, but in practice this can be efficiently represented with a sparse tensor.
The inter-domain scheme has the advantage that it can be used for datasets with a single graph or multiple graphs, whereas the intra-domain scheme is not expected to work when we change the graph batch (in the multi-graph case).
\FloatBarrier
\section{Graph Convolutional NNGP Kernel Derivations}\label{app:kernelderiv}
We show how to derive the kernel/covariance expressions given in Eq.~\eqref{eq:sparsegcnngp} (for the sparse inducing point case). Eq.~\eqref{eq:gcnngp_kernelmatrix} (for the full-rank case) is the same, but with inducing points dropped.

We take the graph nodes to have adjacency information as described in Section~\ref{sec:ind_point_schemes} and Appendix~\ref{app:sec:inducing_point_appendix}. In a graph convolutional NNGP, pre-activation features $\mathbf F^\ell_\text{i}, \mathbf F^\ell_\text{t}$ for the inducing and test/train blocks can be written in terms of the post-activations $\Hi^{\ell-1}$, $\Ht^{\ell-1}$,
\begin{align}
\begin{pmatrix}
    \Fi^\ell\\
    \Ft^\ell
\end{pmatrix} &=
\begin{pmatrix}
\Aii & \Ait\\
\Ati & \Att
\end{pmatrix}
\begin{pmatrix}
    \Hi^{\ell-1}\\
    \Ht^{\ell-1}
\end{pmatrix} \v W,
\end{align}
where $\v W\in\mathbb{R}^{N_{\ell-1}\times N_\ell}$ has i.i.d. Gaussian elements with mean zero and variance $1/N_{\ell-1}$.

Remember that each column $\lambda$ of $\v F^\ell$ is i.i.d., so the covariance of $\v F^\ell$ is,
\begin{subequations}
\begin{align}
    \begin{pmatrix}
        \Kii & \Kit\\
        \Kti & \Ktt
    \end{pmatrix} &= \mathbb{E}\bigg[
\begin{pmatrix}
    \v f^{\text{i};\ell}_\lambda\\
    \v f^{\text{t};\ell}_\lambda
\end{pmatrix}
\begin{pmatrix}
    \v f^{\text{i};\ell}_\lambda\\
    \v f^{\text{t};\ell}_\lambda
\end{pmatrix}^T
\bigg]
\\
&=
\begin{pmatrix}
\Aii & \Ait\\
\Ati & \Att
\end{pmatrix}
\begin{pmatrix}
    \Hi^{\ell-1}\\
    \Ht^{\ell-1}
\end{pmatrix} \mathbb{E}[\v w_\lambda \v w^T_\lambda]
\begin{pmatrix}
    \Hi^{\ell-1}\\
    \Ht^{\ell-1}
\end{pmatrix}^T
\begin{pmatrix}
\Aii & \Ait\\
\Ati & \Att
\end{pmatrix}^T\\
&= \begin{pmatrix}
\Aii & \Ait\\
\Ati & \Att
\end{pmatrix}
\left[\tfrac{1}{N_{\ell-1}}\begin{pmatrix}
    \Hi^{\ell-1}\\
    \Ht^{\ell-1}
\end{pmatrix}
\begin{pmatrix}
    \Hi^{\ell-1}\\
    \Ht^{\ell-1}
\end{pmatrix}^T\right]
\begin{pmatrix}
\Aii & \Ait\\
\Ati & \Att
\end{pmatrix}^T.
\end{align}
\end{subequations}
This is the general form of the graph convolutional kernel. However, if the activation function is ReLU, i.e. $\v H^\ell = \text{ReLU}(\v F^\ell)$, then in the limit $N_{\ell-1}\rightarrow\infty$, $N_{\ell-1}^{-1}\v H^{\ell-1} (\v H^{\ell-1})^T \rightarrow \v \Phi(\v K^{\ell-1})$,
where $\v \Phi(\cdot)$ is arccos kernel \citep{cho2009kernel} and $\v K^{\ell-1}$ is the previous layer's kernel. This gives the recursive relation
\begin{align}
    \begin{pmatrix}
        \Kii^\ell & \Kit^\ell\\
        \Kti^\ell & \Ktt^\ell
    \end{pmatrix}
= \begin{pmatrix}
\Aii & \Ait\\
\Ati & \Att
\end{pmatrix}
{\v \Phi}\left(
    \begin{pmatrix}
        \Kii^{\ell-1} & \Kit^{\ell-1}\\
        \Kti^{\ell-1} & \Ktt^{\ell-1}
    \end{pmatrix}
    \right)
\begin{pmatrix}
\Aii & \Ait\\
\Ati & \Att
\end{pmatrix}^T.
\end{align}
We call this kernel the graph convolutional kernel.
\subsection{Sparse Graph Convolutional Deep Kernel Machine Algorithm}\label{sec:nystrom}
For completeness, we provide a Algorithm~\ref{alg:pred} to demonstrate node prediction with the sparse graph convolutional DKM. We discuss how to parameterize the learned inducing points for training in Appendix~\ref{app:parameterization}. The algorithm can be modified for graph classification by adding a mean-pool layer immediately before sampling.

For the `sample' step at the top-layer, we have two main options. The first, most conventional method, is to sample with a sparse GP (similar to~\cite{milsom2023cdkm}). The second, which is the approach taken in this paper, is to sample weights $\v W$ from the approximate posterior (Eq.~\ref{eq:top_layer_approx_posterior}), which in turn allows sampling logits by calculating ${\Kti \text{chol}(\Kii)^{-1}\v W\in\mathbb{R}^{\ppt \times \nu_{L+1}}}$.
\begin{algorithm}
\caption{Graph convolutional DKM node classification}
  \label{alg:pred}

\let\AND\classAN
\let\algoAND\AND
\begin{algorithmic}
  \STATE {\bfseries Parameters:} $\{\nu_\ell\}_{\ell=1}^L$
  \STATE {\bfseries  Train/test inputs, train labels:} ${\v X}_\mathrm{t}$, $\v Y_\mathrm{t}$
  \STATE {\bfseries Inducing and train/test adjacencies:} $\Aii$, $\Ati$, $\Att$
  \STATE {\bfseries Inducing inputs and inducing Gram matrices:} $\v X_\mathrm{i},\,\{{\v G}^\ell_\mathrm{ii}\}_{\ell=1}^{L+1}$
  \STATE {\bfseries Variational parameters for the output layer:} $\boldsymbol \mu_1,\dots,\boldsymbol \mu_{\nu_{L+1}}, \boldsymbol \Sigma$
  \STATE \textcolor{gray}{Initialize full Gram matrix}
  \STATE $\begin{pmatrix}
    {\v G}^0_\mathrm{ii} & {\v G}^{0}_\mathrm{it} \\
    {\v G}^0_\mathrm{ti} & {\v G}^0_\mathrm{tt}
  \end{pmatrix} \gets \frac{1}{\nu_0}
  \begin{pmatrix}
    {\v X}_\mathrm{i} {\v X}_\mathrm{i}^T & {\v X}_\mathrm{i} {\v X}_\mathrm{t}^T\\
    {\v X}_\mathrm{t} {\v X}_\mathrm{i}^T & {\v X}_\mathrm{t} {\v X}_\mathrm{t}^T
  \end{pmatrix}$
  \FOR{$\ell$ {\bfseries in} $(1,\dotsc,L)$}
    \STATE \textcolor{gray}{Apply kernel non-linearity $\v \Phi$ and perform graph convolution}
    \STATE $\begin{pmatrix}
      {\v K}_\mathrm{ii} & {\v K}_\mathrm{it} \\
      {\v K}_\mathrm{ti} & {\v K}_\mathrm{tt}
    \end{pmatrix} \gets
    \begin{pmatrix}
      {\v A}_\mathrm{ii} & {\v A}_\mathrm{it} \\
      {\v A}_\mathrm{ti} & {\v A}_\mathrm{tt}
    \end{pmatrix}
        {\v \Phi}\left({
    \begin{pmatrix}
      {\v G}^{\ell-1}_\mathrm{ii} & {\v G}^{\ell-1}_\mathrm{it} \\
      {\v G}^{\ell-1}_\mathrm{ti} & {\v G}^{\ell-1}_\mathrm{tt}
    \end{pmatrix}}\right)
    \begin{pmatrix}
      {\v A}_\mathrm{ii} & {\v A}_\mathrm{it} \\
      {\v A}_\mathrm{ti} & {\v A}_\mathrm{tt}
    \end{pmatrix}^T$
    \STATE \textcolor{gray}{Propagate the test-test and test-inducing blocks}
    \STATE ${\v G}_\mathrm{ti}^\ell \gets {\v K}_\mathrm{ti} {\v K}_\mathrm{ii}^{-1} {\v G}^\ell_\mathrm{ii}$
    \STATE ${\v G}_\mathrm{tt}^\ell \gets {\v K}_\mathrm{tt} - {\v K}_\mathrm{ti} {\v K}_\mathrm{ii}^{-1} {\v K}_\mathrm{it} + {\v K}_\mathrm{ti} {\v K}_\mathrm{ii}^{-1} {\v G}^\ell_\mathrm{ii} {\v K}_\mathrm{ii}^{-1} {\v K}_\mathrm{it}$
  \ENDFOR
  \STATE \textcolor{gray}{Calculate output kernel}
    \STATE $\begin{pmatrix}
      {\v K}_\mathrm{ii} & {\v K}_\mathrm{it} \\
      {\v K}_\mathrm{ti} & {\v K}_\mathrm{tt}
    \end{pmatrix} \gets {\v \Phi}{\left(
    \begin{pmatrix}
      {\v G}^{L}_\mathrm{ii} & {\v G}^{L}_\mathrm{it} \\
      {\v G}^{L}_\mathrm{ti} & {\v G}^{L}_\mathrm{tt}
    \end{pmatrix}\right)}$
    \STATE \textcolor{gray}{Perform prediction using a sparse GP (or alternative model) with output kernel}
    \STATE $\v{\hat{ Y}}\gets \text{ sample } \P(\v Y \mid \Kii, \Kti,\v K_\mathrm{tt})$
\end{algorithmic}
\end{algorithm}
\let\classAN\AND
\FloatBarrier
\section{Closed-form Solution for a Linear Graph Convolutional Deep Kernel Machine}\label{app:closed_form_linear}

For some node regression problems with a linear kernel, $\v K(\v G) = \v G$, we are able to find an optimum for the DKM analytically. Assuming a constant regularizer coefficient for each layer, $\nu_\ell = 1$, and no observation noise, the objective $\mathcal{L}$ for a linear graph convolutional DKM is,
\begin{subequations}\label{eq:lineargcdkmobj}
\begin{align}
	\mathcal{L}(\v G^{1},\ldots, \v G^{L}) &= - \sum_{\ell=1}^{L+1}\KL(\mathcal{N}(\v 0, \v G^\ell)\mid\mid \mathcal{N}(\v 0, \hA \v K(\v G^{\ell-1})\hA^T))\\
&= \const\ +  \frac{\nu}{2}\sum_{\ell=1}^{L+1} \left[\log \abs{(\hat{\v A}\v G^{\ell-1}\hat{\v A}^T)^{-1} \v G^{\ell }} - \mathrm{Tr}((\hat{\v A}\v G^{\ell-1}\hat{\v A}^T)^{-1}\v G^{\ell })\right].
\end{align}
\end{subequations}
A slight abuse of notation follows,  in that superscripts $\ell$ on $\v G^\ell$ and $\v Z^\ell$ refer to values at the $\ell$'th layer, but elsewhere, superscripts for example on $\hA^\ell$ refer to powers of $\hA$, so that $\hA^\ell = \hA \cdots \hA$.

For objective Eq.~\eqref{eq:lineargcdkmobj} to make sense, we must assume that the normalized adjacency matrix, $\hat{\v A}$, is invertible. In practice, $\hat{\v A}$ may not be invertible, but we can resolve this by interpolating it with the identity,
\begin{align}
\hA_\lambda = \lambda \v I +  (1-\lambda)\hA,\,\lambda\in(0, 1).
\end{align}
With this adjustment to the adjacency matrix, we can simplify the objective:
\begin{equation}
	\mathcal{L}(\v G^{1},\ldots, \v G^{L}) = \const\
 -\frac{\nu}{2}\sum_{\ell=1}^{L+1} \mathrm{Tr}((\hat{\v A}\v G^{\ell-1}\hat{\v A}^T)^{-1}\v G^{\ell }).
\end{equation}
From here, we drop the transposes because $\hA$ and $\v G^\ell$ are symmetric.
We take gradients:
\begin{subequations}
\begin{align}
	\frac{\partial\mathcal{L}(\v G^{1},\ldots, \v G^{L})}{\partial \v G^\ell} &=-\frac{\nu}{2} \frac{\partial}{\partial\v G^\ell}\left[
 \mathrm{Tr}((\hat{\v A}\v G^{\ell-1}\hat{\v A})^{-1}\v G^{\ell } +
  \mathrm{Tr}((\hat{\v A}\v G^{\ell}\hat{\v A})^{-1}\v G^{\ell+1}
 )\right]\\
 &= \frac{\nu}{2} \left[
-(\hA\G^{\ell-1}\hA)^{-T}  + ( (\v G^\ell)^{-1} \hA^{-1} \v G^{\ell+1}\hA^{-1} (\v G^\ell)^{-1})^T
 \right].
\end{align}
\end{subequations}
Then set those gradients to zero:
\begin{align}\label{eq:AGrel}
\implies \v G^\ell\hA^{-1} (\v G^{\ell-1})^{-1}\hA^{-1} = \hA^{-1} \v G^{\ell+1}\hA^{-1}(\G^\ell)^{-1}.\end{align}
For convenience, define, ${\v Z}^\ell = \v G^\ell \hA^{-1} (\v G^{\ell-1})^{-1}$, then,
\begin{subequations}
\begin{align}
\text{Eq. }\ref{eq:AGrel} \implies  {\v Z}^2 &= \hA {\v Z}^1 \hA^{-1}\\
{\v Z}^3 &= \hA {\v Z}^2 \hA^{-1} = \hA^2 {\v Z}^1 \hA ^{-2}\\
&\vdots\\
{\v Z}^\ell &= \hA^{\ell-1} {\v Z}^1 \hA ^{-\ell+1}.\label{eq:Yell_relation}
\end{align}
\end{subequations}
By the definition of ${\v Z}^\ell$, we have that,
\begin{subequations}
\begin{align}
\prod_{\ell'=1}^\ell{\v Z}^{\ell'} &= (\v G^\ell \hA^{-1}(\v G^{\ell-1})^{-1})(\v G^{\ell-1} \hA^{-1} (\v G^{\ell-2})^{-1})\cdots((\v G^2\hA^{-1} (\v G^1)^{-1})(\v G^1\hA^{-1}(\v G^0)^{-1})\\
&= {\v G}^\ell \hA^{-\ell} ({\v G}^0)^{-1}\label{eq:prodY1}.
\end{align}
\end{subequations}
Also, by Eq.~\eqref{eq:Yell_relation}, we have
\begin{subequations}
\begin{align}
\prod_{\ell'=1}^\ell{\v Z}^{\ell'} &= (\hA^{\ell-1}{\v Z}^1 \hA^{-\ell+1})(\hA^{\ell-2}{\v Z}^1 \hA^{-\ell+2})\cdots(\hA^{1}{\v Z}^1 \hA^{-1})(\hA^{0}{\v Z}^1 \hA^{0})\\
&= \hA^{\ell-1}({\v Z}^1 \hA^{-1})^\ell \hA.\label{eq:prodY2}
\end{align}
\end{subequations}
By setting $\ell=L+1$, and combining equations~\ref{eq:prodY1} and~\ref{eq:prodY2}, we can solve for ${\v Z}^1$,
\begin{subequations}
\begin{align}
{\v G}^{L+1}\hA^{-(L+1)}({\v G}^0)^{-1} &= \hA^{L} ({\v Z}^1 \hA^{-1})^{L+1}\hA\\
\implies {\v Z}^1 & = (\hA^{-L}{\v G}^{L+1}\hA^{-(L+1)}(\v G^0)^{-1}\hA^{-1})^{1/(L+1)} \hA.
\end{align}
\end{subequations}
Matrix roots are not unique, but we calculate roots by performing an eigendecomposition.
Finally we use this expression for ${\v Z}^1$ to obtain a relationship between the kernel representation at layer $\ell$, $\v G^\ell$, and the input and label kernels, ${\v G}^0$ and ${\v G}^{L+1}$,
\begin{align}\label{eq:final_linear_result}
{\v G}^\ell 
 &= \hA^{\ell-1}((\hA^{-L}{\v G}^{L+1}\hA^{-L})(\hA{\v G}^0\hA)^{-1})^{\ell/(L+1)} (\hA{\v G}^0 \hA)\hA^{\ell-1}.
\end{align}
\subsection{Validity of the Solution}
Note that the solution given by Eq.~\eqref{eq:final_linear_result} is only valid if ${\v G}_\ell$ is symmetric and positive definite. The latter property, positive definiteness, can be shown by the fact that ${\v G}_\ell$ is a product of positive definite matrices. We show symmetricity with the following argument.

Let ${\v C =(\hA^{-L}{\v G}^{L+1}\hA^{-L})(\hA{\v G}^0\hA)^{-1}}$. Since $\v C$ is the product of two symmetric PSD matrices, it is PSD (though not necessarily symmetric). Therefore we write $\v C$ using its eigendecomposition, $\v C = \v V \v D \v V^{-1}$, where $\v D$ is the diagonal matrix of the eigenvalues of $\v C$. In particular, $\v D \geq 0$. We write ${\v G}^{L+1}$ in terms of $\v V$ and $\v D$, and use the fact that ${{\v G}}^{L+1}$ is a kernel matrix,
\begin{subequations}
\begin{align}
{\v G}^{L+1} &= \hA^L \v V \v D \v V^{-1} (\hA {\v G}^0 \hA) \hA^{L}\\
&= \hA^L (\hA {\v G}^0 \hA) \v V^{-T} \v D \v V^T \hA^L = ({\v G}^{L+1})^T\\
\implies \v V\v D\v V^{-1} &=  (\hA {\v G}^0 \hA) \v V^{-T} \v D \v V^T (\hA {\v G}^0 \hA)^{-1}.
\end{align}
\end{subequations}
This allows us to show that the powers of $\v C$ have the following property,
\begin{align}
\implies \v C^n = (\v V\v D\v V^{-1})^n =\v V\v D^n\v V^{-1}  =  (\hA {\v G}^0 \hA) \v V^{-T} \v D^n \v V^T (\hA {\v G}^0 \hA)^{-1}.
\end{align}
In particular, we have
\begin{align}
\v C^{\ell/(L+1)} = (\hA {\v G}^0 \hA) \v V^{-T} \v D^{\ell/(L+1)} \v V^T (\hA {\v G}^0 \hA)^{-1}.
\end{align}
Therefore,
\begin{align}
{\v G}^\ell &= \hA^{\ell-1}\v C^{\ell/(L+1)} \hA{\v G}^0 \hA^\ell\\
&= \hA^{\ell-1}\v V \v D^{\ell/(L+1)} \v V^{-1}(\hA{\v G}^0 \hA)\hA^{\ell-1}\\
&= \hA^{\ell-1}\v (\hA {\v G}^0 \hA) \v V^{-T} \v D^{\ell/(L+1)} \v V^T  \hA^{\ell-1}\\
&= ({\v G}^\ell)^T
\end{align}
as desired.

\subsection{Experimental Confirmation of Linear Analytic Results}\label{app:experimental_form_linear_confirm}
\begin{figure*}[!t]
    \centering
    \includegraphics[scale=0.8]{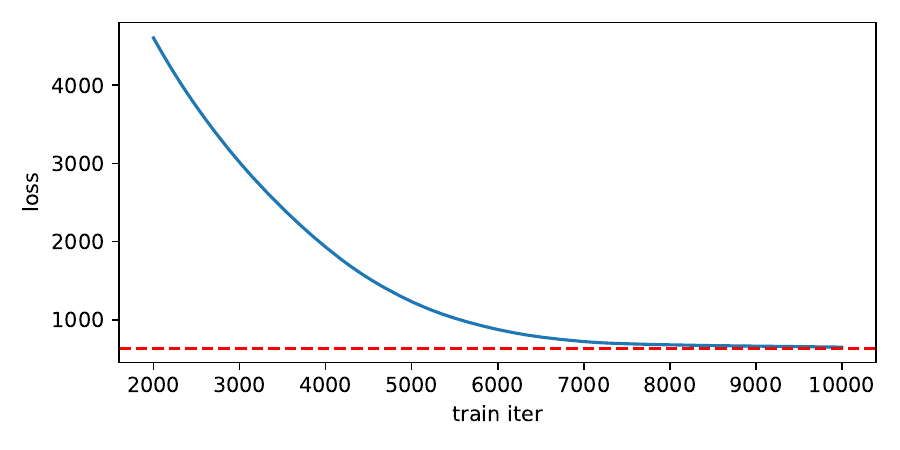}
    \caption{Losses during SGD training of the linear DKM. The loss of the analytic solution is shown by the dashed red line.}
    \label{fig:linear_confirm_plot}
\end{figure*}

Figures~\ref{fig:linear_confirm_kernel} and~\ref{fig:linear_confirm_plot} demonstrate an exact match between our closed-form solution (Eq.~\ref{eq:final_linear_result}) and when optimizing with gradient descent. We selected a random 200 node subset of Cora and (a) calculated the Gram matrices in closed-form, and (b) optimized a linear 2-layer graph convolutional DKM with Adam for 10000 epochs. We used a polynomial learning rate schedule, initialized at 0.1 and decaying with power $0.7$.

Figure~\ref{fig:linear_confirm_kernel} visualizes the Gram matrix representations themselves, and we can see an exact match between the closed-form and gradient descent solutions. Figure~\ref{fig:linear_confirm_plot} shows that the training loss of the gradient descent solution approaches the loss of the analytic solution (that is, the value of the linear graph convolutional DKM objective when plugging in the analytic solution).
\FloatBarrier
\section{Experimental Details}\label{app:experimental_details}
For final performance metrics, we benchmarked the graph convolutional DKM on 14 node classification datasets and 5 graph classification datasets. We also benchmarked a graph convolutional NNGP and GCN on the same datasets to obtain a suitable baselines. We describe details of the datasets and the experimental method below.

\subsection{Dataset Descriptions}\label{app:qualitative_dataset_detail}
Here, we describe the datasets used in experiments qualitatively.
The majority of the datasets we used for benchmarking were node classification datasets, where the task is to predict the labels of all nodes in a single graph, given labels for only a limited number of nodes.~\cora,~\citeseer, and~\pubmed~(Planetoid datasets) are all citation networks, where nodes represent scientific publications, and edges are citations between those publications; one-hot keyword encodings are provided as features, and the goal is to place documents into predefined categories.~\chameleon~and~\squirrel~are both Wikipedia networks on their respective topics, so that nodes correspond to articles, and edges correspond to links between articles; features are again one-hot as in the citation networks, but represent the presence of relevant nouns, and the task is to categories pages based on their average daily traffic.~\penn~\citep{lim2021large} is an online social network where nodes represent users, and edges represent social connections. The task is to predict the reported gender of the users. \flickr~is a social network where nodes represent users, and edges correspond to friendships among users. The labels represent the interest groups of the users.~\blogcatalog~is a network dataset with social relationships of bloggers from the BlogCatalog website, where nodes’ attributes are constructed by the keywords of user profiles. The labels represent the topic categories provided by the authors.
Several of the remaining datasets were introduced by~\citet{platonov2023critical}:
\begin{itemize}
\item\romanempire~is derived entirely from the Roman Empire article on Wikipedia. Nodes are words in the text, and the words are `connected' if they are either adjacent in a sentence or if they are dependent on each other syntactically. Features are word embeddings, and the task is to label each word's syntactic role.
\item\tolokers~is a dataset of workers from the Toloka platform. The nodes are workers, and nodes are connected if workers have worked on the same task. Features are statistics about each worker, and the task is to predict if a worker has been banned.
\item\minesweeper~is a synthetic dataset. The graph has 10,000 nodes corresponding to the cells in the 100 by 100 grid in the Minesweeper game. Nodes are connected if the associated cells are adjacent. Features are one-hot encoded numbers of neighbouring mines, plus a extra feature for when the number of neighbouring mines is unknown. The task is to predict which nodes are mines.
\item\amazonratings~is a graph where nodes are Amazon products, and nodes are connected if they are commonly purchased together. Features are text embeddings of the product descriptions. The task is to predict the average user rating.
\end{itemize}

The two remaining node classification datasets,~\reddit~and~\arxiv, are the largest (roughly one order of magnitude larger than the next biggest). In the~\reddit~dataset~\citep{hamilton2017inductive}, the nodes are posts, the features are GloVe embeddings of the posts, the labels represent subreddits, and posts are connected if a user comments on both posts.

The~\arxiv~dataset~\citep{hu2020open} is another citation network; nodes are computer science Arxiv papers, and papers are connected if one cites the other. Node features are Word2Vec embeddings of a paper's title and abstract, and the labels are subject areas.

We also perform experiments with graph classification datasets. The graph
classification datasets all contain molecules. NCI1 and NCI109 both contain
graphs of chemical compounds (nodes representing atoms, with nodes connected
if there are chemical bonds between them), where the goal is to predict the effect
of the compounds versus lung and ovarian cancer.
Proteins is a dataset of proteins molecules, where the task is to classify as enzymes or
non-enzymes. Mutag is a dataset of compounds, and the task is to predict
mutagenicity on Salmonella typhimurium. Finally, Mutagenicity is a dataset of
drug compounds, and the task is to categorize into mutagen or non-mutagen.
\subsection{Details of Training, Hyperparameter Selection and Final Benchmarking}\label{app:train_details}
\subsubsection{Node classification datasets}\label{sec:ncd_hyp}
For node classification datasets, we used train/validation/test splits from the \texttt{torch\_geometric} library~\citep{Fey/Lenssen/2019} for most datasets; the exceptions were~\arxiv~and~\reddit~for which we adapted code from~\cite{niu2023graph}.
For data preprocessing, we (a) normalized all adjacency matrices as in~\citet{kipf2016semi},  and (b) we scaled input features such that the associated linear kernel had entries in $[-1,1]$, i.e.
\begin{align}
X_{i\lambda}' = \frac{X_{i\lambda}}{\sum_{\mu}X_{i\mu}{X_{i\mu}}}.
\end{align}
We found that this transformation helped avoid numerical errors when training the models, as it ensures $\v G^0 = \v X' (\v X')^T$ has elements all on roughly the same scale.

To select graph DKM hyperparameters, we started with a base model analogous to the GCN described by~\cite{kipf2016semi}, but with an extra Gram layer before the first kernel non-linearity.
From this base model, we performed 4 sweeps to select (1)~\dof~and inducing point scheme, (2) architecture, (3) centering parameters, and finally (4) the number of inducing points.
The sweeps were,
\begin{itemize}
\item Sweep (1): $\nu\in\{0, 0.01, 0.1, 1, 10, 100, 1000, \infty\}$, $\verb|scheme|\in\{\text{inter},\, \text{intra}\}$,
\item Sweep (2): $\verb|arch|\in\{\verb|kipf|,\, \verb|kipf|+\hA_{\lambda},\, \verb|kipf|+\hA_{\lambda}+\verb|resid|,\, \verb|platinov|\}$. Here, \verb|kipf| refers to the default architecture, $+\hA_{\lambda}$ means that $\hA_\lambda$ was used instead of the usual renormalized adjacency matrix, +\verb|resid| refers to the addition of residual blocks, and \verb|platinov| refers to an architecture similar to the one used in~\cite{platonov2023critical}. The \verb|platinov| architecture is similar to a residual 2-layer GCN, but with two linear layers in each residual block.
\item Sweep (3) searched over centering or no centering, and whether to include a learned scale and bias parameter (described in Appendix~\ref{app:kernel_centering}).
\item Sweep (4): $\ppi\in\{50, 100, 200, 300, 400\}$.
\end{itemize}

We trained for 300/200 epochs in sweep (1), and in the remaining sweeps we trained for  200/150 for the smaller/bigger datasets (where~\arxiv~and~\reddit~and the `bigger' datasets). For the smaller datasets `one epoch' means a single full-batch gradient descent step; for the bigger datasets it means two mini-batched gradient descent steps.

We had a numerical stability issue with~\squirrel~when selecting hyperparameters. To resolve this, we set \verb|arch=kipf|, rather than using \verb|arch=platinov|, which was the `best' according to sweep (2).

We repeated the above sweeps, but with DKM regularization fixed to $\nu=\infty$ (equivalent to removing Gram layers) to obtain final performance metrics for a sparse graph convolutional NNGP. To find suitable hyperparameters for the GCN, we used a single grid search starting with the base GCN model of~\cite{kipf2016semi}. We searched over $\verb|width|\in\{100, 200\}$, \verb|dropout| $\in\{0, 0.5\}$, \verb|batchnorm| $\in\{\text{yes}, \text{no}\}$, and \verb|row_normalization| $\in\{\text{yes}, \text{no}\}$, $\verb|arch|\in\{\verb|kipf|,\,\verb|platinov|\}$.

We used the Adam optimizer with a two-stage learning rate schedule for all training runs. We increase the learning rate linearly from $10^{-3}$ to $10^{-2}$ for the first quarter of the epochs, and after that use a cosine schedule with a minimum learning rate of $10^{-5}$.
The models were written in Pytorch, and we trained on a cluster containing RTX 2080's, RTX 3090's and A100s.

\subsubsection{Graph classification datasets}
 For graph classification datasets, we constructed our own cross-validation splits (10 splits), since no default split was provided by \texttt{torch\_geometric}. We adopted a very similar approach for hyperparameter selection as in Section~\ref{sec:ncd_hyp}. The main differences being that we set \verb|scheme|$=$inter (since an intra-domain inducing-point scheme is not applicable in the multi-graph setting), and we trained for 300 batches on all datasets.

 To perform graph classification with node-level features/kernels of multiple graphs, we (a) stack graphs together in batches of 1024, and (b) perform a mean aggregation over nodes in each graph at the top layer of the network, thus obtaining logits for each graph.
\subsection{Dataset Statistics}
We include dataset statistics for node classification datasets in Table~\ref{tab:node_classification_statistics} and Table~\ref{tab:graph_classification_statistics} respectively.
\begin{table*}[!t]
  \centering
  \caption{Node classification dataset statistics.} \label{tab:node_classification_statistics}
  \vskip 0.15in
    \begin{tabular}{cccccc}
        \toprule
        Dataset & \# nodes & \# edges & Homophily ratio & \# features & \# classes \\
        \midrule
        \romanempire & 22,662 & 32,927   & 0.05  & 300 & 18 \\
        \squirrel & 5,201 & 198,353 & 0.22 & 2,089 &  5 \\
        \flickr & 7,575 & 239,738 &  0.24 & 12,047 & 9 \\
        \chameleon & 2,227 & 31,371 & 0.23 & 2,325 &  5 \\
        \amazonratings & 24,492 & 93,050 & 0.38 & 300 &  5 \\
        \blogcatalog & 5,196 & 171,743 & 0.40 & 8,189 & 6\\
        \penn & 41,554 & 1,362,229 & 0.47 & 4,814 & 2\\
        \tolokers & 11,758 &  519,000 & 0.59  & 10 & 2  \\
        \arxiv & 169,343 & 1,157,799 & 0.65 & 128 & 40 \\
        \minesweeper & 10,000 & 39,402 & 0.68 & 7 &  2 \\
        \citeseer &  3,327 & 4,552 & 0.74 & 3,703 & 6 \\
        \reddit & 232,965 & 57,307,946 & 0.76 & 602 & 41 \\
        \pubmed & 19,717 & 44,324 & 0.80 & 500 & 3 \\
        \cora &  2,708 & 5,278 & 0.81 & 1,433 & 7 \\
        \bottomrule
    \end{tabular}
\end{table*}
\begin{table*}[htbp]
  \centering
  \caption{Graph classification dataset statistics.} \label{tab:graph_classification_statistics}
  \vskip 0.15in
    \begin{tabular}{cccccc}
        \toprule
        Dataset & \# graphs & avg. \# nodes & avg. \# edges & \# features & \# classes \\
        \midrule
        Proteins & 1,113 & 39.1 & 72.8 & 5 & 2 \\
        NCI1 & 4,110 & 29.9 & 32.3 & 38 & 2 \\
        NCI109 & 4,127 & 29.7 & 32.1 & 39 & 2 \\
        Mutag & 188 & 17.9 & 19.8 & 8 & 2 \\
        Mutagenicity & 4,337 & 30.3 & 30.8 & 15 & 2 \\
        \bottomrule
    \end{tabular}
\end{table*}
\subsection{Details of the Other Experiments}
For Figure~\ref{fig:linear_confirm_kernel} and Figure~\ref{fig:linear_confirm_plot}, we trained a linear graph convolutional DKM for 10,000 epochs on a randomly selected 200 node subset of~\cora.

For Figure~\ref{fig:analyze_linear}, we calculate the linear graph convolutional DKM kernels on randomly generated~\erdosrenyi~graphs, with 50 nodes and edge probability 0.1. When calculating the kernels, we use $\hA_{\lambda=0.5}$ for the graph mixup. The input kernels are standard Wishart samples with 50 degrees of freedom. The nodes are randomly assigned one of two classes, with an even split.

For Figure~\ref{fig:shaped} we trained graph convolutional DKMs with the arc-cosine kernel non-linearity for 300 epochs, with regularization~$\nu\in\{0, 1, 10, 1000\}$, and $\hA_{\lambda=0.3}$ for graph mixup. The kernel alignment (CKA) figures were computed using the full kernels, without any normalization.

For all of the Figures described above in this subsection, we used a simple 2-layer architecture, similar to~\cite{kipf2016semi}. All kernel plots are normalized to ensure entries are between $[-1,1]$.

Finally, to produce Figure~\ref{fig:val_acc_vs_dof_kipf} we used data from the first hyperparameter sweep.
\section{Kernel Centering Layer}\label{app:kernel_centering}
As part of our grid search, we considered adding a `kernel centering layer'. When enabled, the centering layer performs a centering similar to batchnorm,
\begin{align}
(\Ft)'_{i\lambda} = (\Ft)_{i\lambda} - \frac{1}{\ppt}\sum_{j}(\Ft)_{j\lambda},
\end{align}
where $\v K = \Ft\Ft^T$ is the kernel to be centered. Additionally, we include the option of learning a scale parameter $\gamma$ and bias parameter $\beta$,
\begin{align}
(\Ft)''_{i\lambda} = \gamma (\Ft)'_{i\lambda} + \beta.
\end{align}
\section{Inducing-point ablations}\label{app:ind_point_abl}
Table~\ref{tab:ind_abl} shows validation accuracies for different inducing point schemes. We obtained these accuracies during Sweep (1), so that $\nu$ has been tuned for each inducing-point scheme (but not other hyperparameters). We find that inter-domain is a better scheme overall, but intra-domain can be beneficial for a few datasets (most notably Penn94).
\begin{table*}[htbp]
  \centering
  \caption{Validation accuracies for the two inducing point scheme on a range of node classification datasets.} \label{tab:ind_abl}
  \vskip 0.15in
\begin{tabular}{lll}
\toprule
 & Inter-domain & Intra-domain \\
\midrule
Cora & $\mathbf{78.9 \pm 0.2}$ & $58.2 \pm 0.5$ \\
Pubmed & $\mathbf{79.4 \pm 0.4}$ & $74.4 \pm 1.6$ \\
Reddit & $94.2 \pm 0.0$ & $\mathbf{94.5 \pm 0.0}$ \\
Citeseer & $\mathbf{70.8 \pm 0.2}$ & $23.4 \pm 0.2$ \\
Minesweeper & $\mathbf{80.2 \pm 0.2}$ & $80.1 \pm 0.3$ \\
Arxiv & $69.6 \pm 0.1$ & $\mathbf{70.2 \pm 0.1}$ \\
Tolokers & $79.9 \pm 0.4$ & $\mathbf{80.5 \pm 0.5}$ \\
Penn94 & $69.5 \pm 0.3$ & $\mathbf{81.3 \pm 0.4}$ \\
BlogCatalog & $\mathbf{76.0 \pm 0.8}$ & $72.8 \pm 1.3$ \\
Amazon Ratings & $46.3 \pm 0.3$ & $\mathbf{47.8 \pm 0.5}$ \\
Flickr & $\mathbf{59.2 \pm 1.6}$ & $55.8 \pm 1.6$ \\
Chameleon & $\mathbf{66.5 \pm 2.0}$ & $66.0 \pm 2.0$ \\
Squirrel & $49.9 \pm 0.7$ & $\mathbf{54.4 \pm 0.9}$ \\
Roman Empire & $51.2 \pm 0.7$ & $\mathbf{51.7 \pm 0.8}$ \\
\bottomrule
\end{tabular}
\end{table*}
\section{Deep Graph Convolutional DKMs}\label{app:deep_dkms}
\begin{figure*}[!t]
    \centering
    \includegraphics[width=\textwidth]{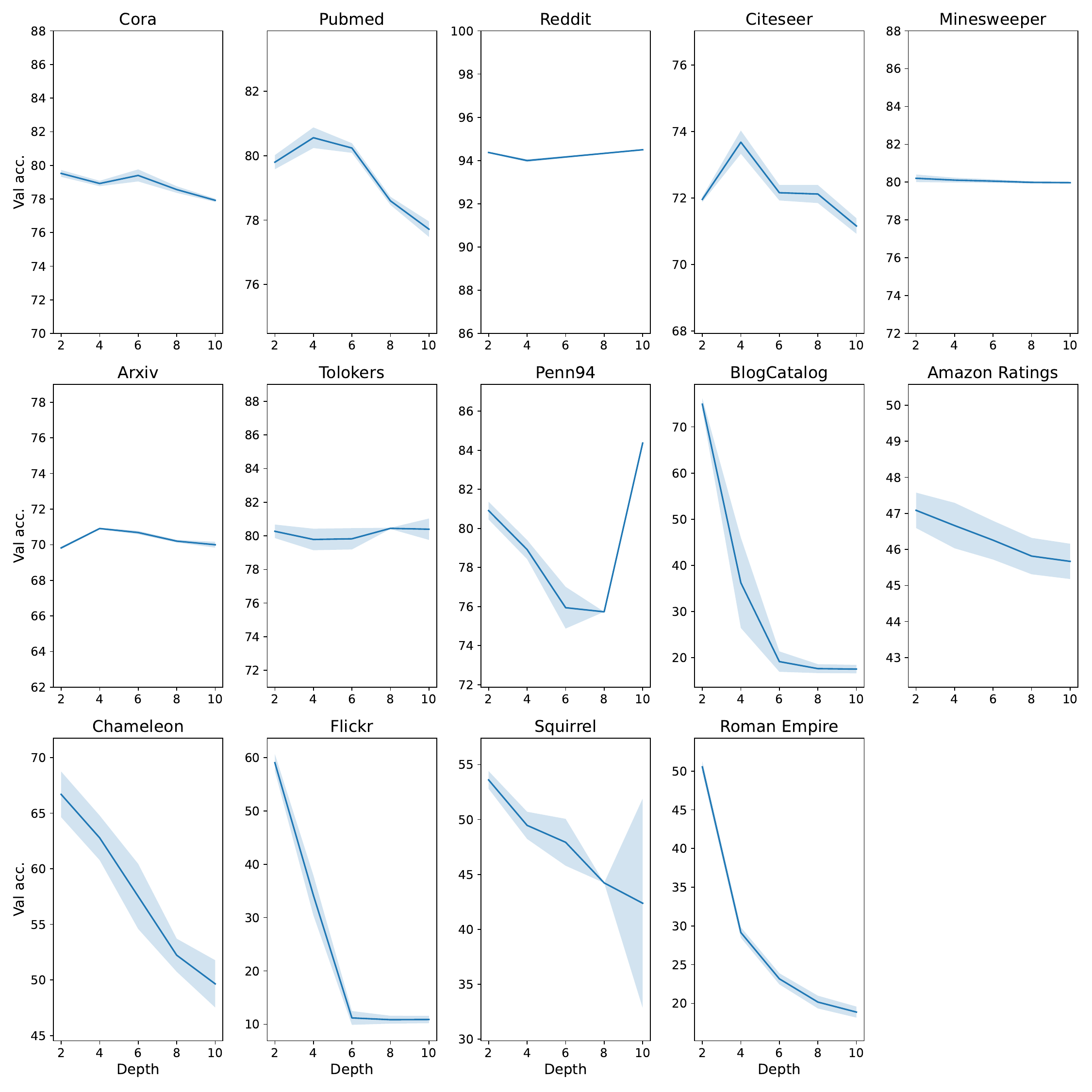}
    \caption{Validation accuracy at different depths for different node classification datasets. Error bands are $\pm$ 1 standard deviation.}
    \label{fig:val_acc_vs_depth}
\end{figure*}

We investigated how the graph convolutional DKM performs as we increase the number of layers, with results shown in Figure~\ref{fig:val_acc_vs_depth}. Similarly to the inducing point ablations (Section~\ref{app:ind_point_abl}), we show validation accuracies using settings found in Sweep (1), so that $\nu$ and inducing-point scheme have been tuned (but not other hyperparameters). We find that increasing depth tends not to help performance (though it can occassionally help in some cases, e.g.~\citeseer). This finding is expected, since GCN also loses performance as depth is scaled~\citep{oono2019graph}. Note that we encountered a numerical instability when running the Reddit depth experiment, so the relevant data points have been omitted (depth=$6$ and $8$).
\FloatBarrier
\section{Parameterization of the Deep Kernel Machine}\label{app:parameterization}
Even though the inducing point scheme significantly reduces the computational complexity of training a DKM, computing the objective and forward equations is still expensive and has the potential to be unstable. In particular, the KL regularization terms
\begin{align}
\KL\big(\normal(\v 0, &\v G^\ell)\mid\mid\normal(\v 0, \v  K(\v G^{\ell-1}))\big) =\nonumber\\
&\frac{1}{2}\big(P-\log\det(\v K(\v G^{\ell-1})^{-1} \v G^{\ell }) + \mathrm{Tr}(\v K(\v G^{\ell-1})^{-1}\v G^{\ell })\big)
\end{align}
involve an unpleasant inverse and $\log$-determinant. When implementing convolutional DKMs,~\cite{milsom2023cdkm} performed these calculations naively, which necessitates the use of double precision to avoid stability issues. The use of double precision is inefficient.

We propose the following parameterization to avoid the $\log\det$, while still computing the regularization term exactly. At each layer, we learn the parameter ${\v L^\ell = \text{cholesky}(\v K(\Gii^{\ell-1})^{-1}\Gii^\ell)}$. Due to the fact that $\v L^\ell$ is lower triangular, the log determinant term can easily be computed as,
\begin{align}
\log\det(\v K(\Gii^{\ell-1})^{-1}\Gii^\ell) =  \log\det(\v L \v L^T) = 2\sum_{j=1}^{\ppi} \log L^\ell_{jj}.
\end{align}
The forward equations given by~\cite{dkm23} are
\begin{subequations}\label{eq:yangforward}
\begin{align}
  \Gti^\ell &= \Kti \Kii^{-1} \Gii^\ell,\\
  \Gtt^\ell &= \Ktt - \Kti \Kii^{-1} \Kit +\Kti \Kii^{-1}\Gii^\ell \Kii^{-1}\Kit.
\end{align}
\end{subequations}
We suggest using the~\nystrom~approximation $\Ktt \approx \Kti \Kii^{-1}\Kit$ to avoid computing ${\Kti \Kii^{-1}\Kit}$. Additionally, we know that there exist $\Fi$ and $\Ft$ such that
\begin{align}
\begin{pmatrix}
\Gii^\ell & \Git^\ell\\
\Gti^\ell & \Gtt^\ell
\end{pmatrix} = \begin{pmatrix}
\Fi\Fi^T & \Fi\Ft^T\\
\Ft\Fi^T & \Ft\Ft^T
\end{pmatrix}.
\end{align}
Therefore the Gram forward equations (Eq.~\ref{eq:yangforward}) with a Nystrom approximation for $\Ktt$ are equivalent to
\begin{subequations}\label{eq:onesolve_scheme}
\begin{align}
\Hi &= \text{cholesky}(\Kii),\\
\Fi &= \Hi \v L^\ell,\\
\Ft &= \Kti \Hi^{-1}\v L^\ell\label{eq:onesolve_Ft}.
\end{align}
\end{subequations}
Notably, the $\Kti\Hi^{-1}$ computation can be achieved using a triangular solve, since $\Hi$ is lower triangular. The trade-off of propagating with Eq.~\eqref{eq:onesolve_scheme} rather than the original scheme (Eq.~\ref{eq:yangforward}) is that in exchange for using the~\nystrom~approximation we get (a) efficient objective computation avoiding $\log\det$, (b) a triangular solve instead of a cholesky solve, and (c) fewer matrix multiplies.
\end{document}